\documentclass[10pt,twocolumn,letterpaper]{article}

\usepackage[pagenumbers]{cvpr} %

\newcommand{\method}{FCA\xspace}
\newcommand{\cmark}{\ensuremath{\checkmark}}
\newcommand{\xmark}{\ensuremath{\times}}

\DeclareMathOperator*{\argmax}{arg\,max}
\newcommand{\Loc}{\mathcal{L}}
\newcommand{\Str}{\mathcal{S}}
\newcommand{\Pre}{\mathcal{P}}

\usepackage{colortbl}
\colorlet{oursrow}{gray!12}
\newcommand{\oursrow}{\rowcolor{oursrow}}

\definecolor{cvprblue}{rgb}{0.21,0.49,0.74}
\usepackage[pagebackref,breaklinks,colorlinks,allcolors=cvprblue]{hyperref}

\def\confName{CVPR}
\def\confYear{2027}

\usepackage{tikz}
\definecolor{fcalocal}{rgb}{0.21,0.49,0.74}
\definecolor{fcastrided}{rgb}{0.80,0.22,0.20}
\DeclareRobustCommand{\fcaicon}{%
  \raisebox{-0.16em}{\begin{tikzpicture}[x=0.25em,y=0.25em]
    \foreach \i in {0,...,3} \foreach \j in {0,...,3} \fill[black!18,rounded corners=0.25pt] (\i+0.1,\j+0.1) rectangle (\i+0.9,\j+0.9);
    \foreach \i in {0,...,3} \fill[fcalocal,rounded corners=0.25pt] (\i+0.1,2.1) rectangle (\i+0.9,2.9);
    \foreach \j in {0,...,3} \fill[fcastrided,rounded corners=0.25pt] (1.1,\j+0.1) rectangle (1.9,\j+0.9);
    \fill[black!75,rounded corners=0.25pt] (1.1,2.1) rectangle (1.9,2.9);
  \end{tikzpicture}}%
}
\title{\vspace{-12mm}\fcaicon\hspace{0.35em}Factorized Context Attention: From Emulated to Native Grid Probing}

\newcommand{\arxivversion}{}   %
\author{%
Mohamed Eltahir$^{1}$ \quad Manar Eyad$^{1,*}$ \quad Yara Alzahrani$^{1,*}$ \quad Fatimah Albesher$^{1,*}$\\
Duaa M. Tahir$^{1}$ \quad Tanveer Hussain$^{2,\dagger}$ \quad Naeemullah Khan$^{1,\ddagger}$\\[3pt]
{\normalsize $^{1}$King Abdullah University of Science and Technology (KAUST), Thuwal, Saudi Arabia}\\
{\normalsize $^{2}$Department of Computer Science, Edge Hill University, Ormskirk, England}\\[2pt]
{\small\texttt{\{mohamed.hamid, manar.ali, yara.zahrani, fatema.besher\}@kaust.edu.sa}}\\
{\small\texttt{\{duaa.tahir, naeemullah.khan\}@kaust.edu.sa, hussaint@edgehill.ac.uk}}
\vspace{-6mm}
}

\begin{document}
\maketitle
\renewcommand{\thetitle}{Factorized Context Attention: From Emulated to Native Grid Probing}   %
{\let\thefootnote\relax\footnotetext{$^{*}$Equal contribution. $^{\dagger}$Corresponding author. $^{\ddagger}$Principal investigator.\\
\hspace*{1.8em}Code: \url{https://github.com/mohammad2012191/FCA}}}
\begin{abstract}
Models read a book, an hour of video, or an image in one forward pass. As the context grows, this pass slows down, loses accuracy, and stops at the context window. Sparse attention is efficient, but keeps the other two limitations. Grid probing, a recent training-free method, recovers the accuracy and extends the reach by asking a frozen model the same question about short local and strided spans of the input, but every question repeats the whole computation, since each span is encoded again. Our insight is that grid probing is emulated attention, an attention pattern reproduced from outside the model through prompts, since each probe only restricts which part of the input the question can see. We introduce \textbf{Factorized Context Attention (\method)}, a training-free block-sparse attention that runs the same pattern natively, as the attention of the model itself. \method encodes the input once into a KV cache, reads all spans as masked sequences in one batched forward, and uses a stair position schedule to keep every span inside the trained position range. Across long video understanding, long-context text inference and image grounding, \method matches the accuracy and the reach of grid probing at a fraction of its cost. It gains about 10 points over the single pass on long video and 9 to 14 points on long novels, finds needles in videos of over a thousand frames and in texts four times the length of the window, and answers a later question about the same input in as little as a tenth of the time of grid probing. Native attention thus keeps what emulated attention gains without its cost, all with existing frozen language and vision-language models.
\end{abstract}

\section{Introduction}
\label{sec:intro}

\begin{figure}[t]
  \centering
  \includegraphics[width=\linewidth]{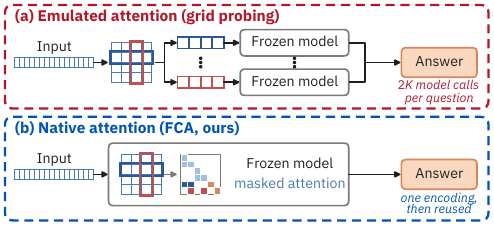}
  \caption{\textbf{Emulated against native attention.} (a) Grid probing builds the grid outside the frozen model and calls the model once per span. (b) \method passes the input once and selects the spans inside the model with masked attention.}
  \label{fig:teaser}
\end{figure}

Models answer questions about a book or a long video by reading the whole input in a single forward pass, which we call the monolithic read. As the input grows, the monolithic read slows down, since its attention cost is quadratic, loses accuracy well before it reaches the context window~\cite{ruler,nolima,lostmiddle,ducontext}, and stops at the window. Sparse attention~\cite{xattention,tokensparse} and cached serving~\cite{kwon2023pagedattention} make the read faster, but neither restores its accuracy or lets it read past the window.

A recent line of training-free methods, which we refer to as grid probing, reads the input differently~\cite{gridprobe,answermap,periscope}. Grid probing places the blocks of the input on a hypothetical $K{\times}K$ grid, asks a frozen model the same question about each row (a local span) and each column (a strided span), and pools the $2K$ answers into one answer that reflects the whole input. Because every span is short, each read stays within the range that the model handles well, which reduces the noise of long contexts and extends the reach beyond the window. Short spans also keep the key-value cache small and manageable while matching or exceeding the accuracy of the monolithic read. This comes, however, at a significant cost in speed and computation, as every span is a separate prompt that is encoded again for every question and is not reused.

Our key observation is that grid probing is attention in disguise. Each probe only decides which span of the input the question may attend to, which is exactly what an attention mask does. We therefore call grid probing \emph{emulated attention}, since it reproduces an attention pattern from outside the model, one prompt at a time, and we call the same pattern run inside the attention layers of the model \emph{native attention}. \Cref{fig:teaser} contrasts the two. Emulated attention needs only the outputs of a model, so it runs on closed models and holds a single span in memory, whereas native attention encodes the input once and reuses that encoding for every span and every question.

We introduce \textbf{Factorized Context Attention} (\method, \cref{fig:fca}), a training-free block-sparse attention pattern that runs grid probing as native attention. \method encodes the input once and stores its keys and values. It then answers each request with $2K$ masked sequences in one batched forward, each attending to one local or strided span, at positions that keep every span contiguous inside the trained range. \method thereby inherits the accuracy, the reach and the evidence map of grid probing, while being faster and cacheable.

Across long video, long text and single images, \method matches the accuracy of grid probing at a fraction of its time, and it gains most over the monolithic read where the input strains the model, by about 10 points on long video and 9 to 14 points on long novels. \method also finds needles in videos of over a thousand frames and in texts four times the window, which the monolithic read cannot reach.

The main contributions of this work are summarized as follows:
\begin{itemize}[leftmargin=*,itemsep=1pt,topsep=2pt]
  \item We introduce the distinction between \textbf{emulated attention}, an attention pattern reproduced from outside a model through prompts, and \textbf{native attention}, the same pattern run inside its attention layers, and we characterize when each is the better choice.
  \item We propose \textbf{Factorized Context Attention}, a training-free block-sparse attention pattern that runs grid probing natively from a single encoding of the input.
  \item We show that \method reproduces the local probes of grid probing up to a shift of positions, and that it keeps the $s^{1.5}$ encoding cost of grid probing while paying it once instead of for every question.
\end{itemize}

\section{Related Work}
\label{sec:related}

A model that reads a long context in a single pass is slow, loses accuracy as the context grows, and stops at its window. Prior work addresses this problem along three lines. The first makes the attention of the pass cheaper, the second avoids encoding the same context twice, and the third avoids reading the whole context at once.

\paragraph{Cheaper attention.}
The first line reduces the cost of attention over a long context, with methods of two categories, (i) trained patterns, which build a sparse pattern into the model, and (ii) training-free methods, which change how a frozen model reads. In the first category, the Sparse Transformer restricts each position to a window of preceding positions and to positions at a fixed interval~\cite{sparsetransformer}, axial attention attends along one axis of a multidimensional input at a time~\cite{axial}, Longformer and BigBird combine sliding windows with global tokens, with random connections in BigBird~\cite{longformer,bigbird}, and recent models route each query to a subset of the context, MoBA and NSA to blocks through a block-scoring gate~\cite{moba,nsa} and DeepSeek-V3.2 to tokens through a learned indexer~\cite{deepseekv32}. All of these patterns are part of the architecture. The model must be trained from scratch or fine-tuned with the pattern, so none of them can be applied post hoc to a frozen model. They also keep a single pass over the whole context, so they still stop at the context window. In the second category, sparse prefills approximate the monolithic read with fewer attention computations~\cite{minference,xattention,tokensparse}, so they keep its window limit and its accuracy loss. Streaming methods read past the window with a sliding window and a memory~\cite{streamingllm,infllm}, but they drop distant context or retrieve it heuristically. Position interpolation stretches the window with little or no fine-tuning~\cite{yarn}, but it still needs one long monolithic read, whose attention cost grows with the square of the context. Efficient kernels are orthogonal to both categories, since they compute the same attention with less memory traffic~\cite{flashattention,flashattention2} or compile arbitrary block masks~\cite{flexattention}.

\begin{table}[t]
  \centering
  \footnotesize
  \setlength{\tabcolsep}{1.6pt}
  \begin{tabular}{@{}lccccc@{}}
    \toprule
    & \textbf{Post} & \textbf{Black-} & \textbf{Encode} & \textbf{Past} & \textbf{Evid.} \\
    \textbf{Method} & \textbf{hoc} & \textbf{box} & \textbf{once} & \textbf{window} & \textbf{map} \\
    \midrule
    Monolithic, prefix cache & \cmark & \cmark & \cmark & \xmark & \xmark \\
    Trained fixed patterns~\cite{sparsetransformer,axial} & \xmark & \xmark & \cmark & \xmark & \xmark \\
    Trained sparse attn.~\cite{moba,nsa} & \xmark & \xmark & \cmark & \xmark & \xmark \\
    Sparse prefill~\cite{xattention,tokensparse} & \cmark & \xmark & \cmark & \xmark & \xmark \\
    Parallel encoding~\cite{pcw,ape} & \cmark & \xmark & \cmark & \cmark & \xmark \\
    Grid probing~\cite{gridprobe,periscope}, emulated & \cmark & \cmark & \xmark & \cmark & \cmark \\
    \midrule
    \oursrow \textbf{\method} (ours), native & \cmark & \xmark & \cmark & \cmark & \cmark \\
    \bottomrule
  \end{tabular}
  \caption{\textbf{Positioning.} Grid probing emulates an attention pattern through prompts, and \method runs the same pattern natively. Post hoc means the method applies to a frozen model with no training. Black-box means the read needs only the model's next-token scores. Encode once means one encoding serves every question on a context. Past window means reading contexts longer than the model's window.}
  \label{tab:positioning}
\end{table}

\paragraph{Encoding context once.}
The second line leaves attention as it is and removes repeated work. Serving systems avoid encoding a shared context again for every question by caching its keys and values~\cite{kwon2023pagedattention} and batching many queries over it~\cite{hydragen}. Parallel encoding goes further. It encodes every context alone and lets one query attend to all of them at once~\cite{pcw,ape,blockattention}. No cached block has seen any other, and accuracy drops unless the cache is repaired, by recomputing tokens~\cite{cacheblend,epic}, by an anchor block~\cite{starattention}, by calibrating the attention~\cite{ape}, or by fine-tuning~\cite{blockattention}. In all of these methods one query still attends to every cached block, so its attention span grows with the context and the quality of the answer depends on how well the cache is repaired. This line saves time on later questions, but each question is still one read over the whole context.

\paragraph{Reading in short spans.}
The third line, grid probing, addresses the accuracy loss and the window limit directly, because it never lets one query read the whole context. Grid probing places the input on a $K{\times}K$ grid, asks a frozen model the same question about each row and each column, and combines the $2K$ answers. GridProbe applies it to the frames of a video and selects frames for a second pass~\cite{gridprobe}, AnswerMap to the bands of an image, to map where an answer comes from~\cite{answermap}, and Periscope to the chunks of a long text, which it reads past the window~\cite{periscope}. All three run at the prompt level, as emulated attention in our terms, so they read any model through its next-token scores and hold the memory of one probe. Their limitation is cost. Every probe is a separate forward pass over its own copy of its span, so the input is encoded twice for every question, no encoding is reused from one question to the next, and every later question costs as much as the first.

The three lines are complementary, and a fast factorized read needs one thing from each, the bounded local and strided spans of grid probing, the single encoding of cached serving, and a block-sparse pattern that a frozen model can run on standard kernels. \Cref{tab:positioning} compares the approaches, and \cref{sec:method} builds \method from these three parts.

\section{Methodology}
\label{sec:method}

An attention pattern decides, for every query token, which keys it may attend to and at which positions. In every attention layer, the output at a query position $t$ is
\begin{equation}
o_t=\sum_{t'\in\mathcal{A}(t)}\operatorname*{softmax}_{t'\in\mathcal{A}(t)}\Bigl(\frac{q_t^{\top}k_{t'}}{\sqrt{d}}\Bigr)\,v_{t'},
\label{eq:attn}
\end{equation}
where $q_t$, $k_{t'}$, and $v_{t'}$ are the layer's query, key, and value vectors, $d$ is the head dimension, and $\mathcal{A}(t)$ is the set of positions that $t$ may attend to. The monolithic read uses the causal set $\mathcal{A}(t)=\{t'\le t\}$ over the whole sequence. Any other pattern can be realized on a frozen model in two ways. \emph{Emulated attention} realizes the pattern from outside the model. It builds one prompt for each set, holding only the tokens that the query may attend to, and runs the model on every prompt with its ordinary causal attention. \emph{Native attention} realizes the pattern inside the model. It keeps one encoding of the input and restricts $\mathcal{A}(t)$ and the positions directly. \Cref{sec:probe} presents grid probing as emulated attention, \cref{sec:pattern} presents \method, which runs the same pattern natively, \cref{sec:stair} gives the positions that the native pattern needs, and \cref{sec:exec} compares the two.

\subsection{Emulated attention: grid probing}
\label{sec:probe}

Let $\theta$ be a frozen decoder-only model and $x$ a context of $s$ tokens, split into $N$ blocks $x_1,\dots,x_N$. For text a block is a chunk of $c$ tokens, and for video it is one frame. We set $K{=}\lceil\sqrt{N}\,\rceil$ and index the blocks as a $K{\times}K$ grid, block $x_{(i-1)K+j}$ at cell $(i,j)$, with cells beyond $N$ left empty. The $i$-th local span and the $j$-th strided span are
\begin{equation}
\begin{aligned}
\Loc_i &= \bigl(x_{(i-1)K+1},\dots,x_{iK}\bigr),\\
\Str_j &= \bigl(x_{j},\,x_{j+K},\dots,x_{j+(K-1)K}\bigr),
\end{aligned}
\label{eq:spans}
\end{equation}
each $K$ blocks in reading order. A local span covers a stretch of consecutive blocks, and a strided span samples the whole context. Every block lies in exactly one local and one strided span.

A request $r$ is the question with its answer options, and $p$ is the fixed prompt that precedes the context. For a span $T$, a probe reads the model's next-token distribution after the sequence $(p,T,r)$. For options $\mathcal{Y}$ and an abstain option $\bar{y}$, the probe score of option $y$ is its log-odds against abstaining~\cite{periscope},
\begin{equation}
\ell(y\mid T)=\log p_\theta(y\mid p,T,r)-\log p_\theta(\bar{y}\mid p,T,r),
\label{eq:probe}
\end{equation}
so a span without evidence scores every option low. We write $L_i(y){=}\ell(y\mid\Loc_i)$ and $S_j(y){=}\ell(y\mid\Str_j)$. The pooled score and the evidence map are
\begin{equation}
\begin{aligned}
\mathrm{score}(y) &= \max_i L_i(y)+\max_j S_j(y),\\
M[i,j] &= \max_{y\in\mathcal{Y}}\bigl(L_i(y)+S_j(y)\bigr).
\end{aligned}
\label{eq:pool}
\end{equation}
An option scores high only when some local span and some strided span both support it, and every block is scored by the two spans it lies in. The \emph{direct} read-out answers $\argmax_y\mathrm{score}(y)$. The \emph{select} read-out reads the $K$ blocks with the highest cells of $M$, in their original order, with the request, and answers or generates from that read, which encodes the chosen blocks again as one short sequence. Open-ended questions use the same read with the options Yes and No, whether the span contains the answer~\cite{periscope}, and the select read-out then generates the answer from the blocks with the highest cells.

Grid probing computes each $\ell(\cdot\mid T)$ with a separate forward pass over the prompt $(p,T,r)$. Inside that prompt the request attends to $T$ and to no other part of the context, and $T$ sits at consecutive positions after $p$, so the $2K$ prompts emulate $2K$ attention sets with the ordinary causal attention of the model. The price is that every probe encodes its span again, for every request.

\begin{figure*}[!t]
  \centering
  \includegraphics[width=\linewidth]{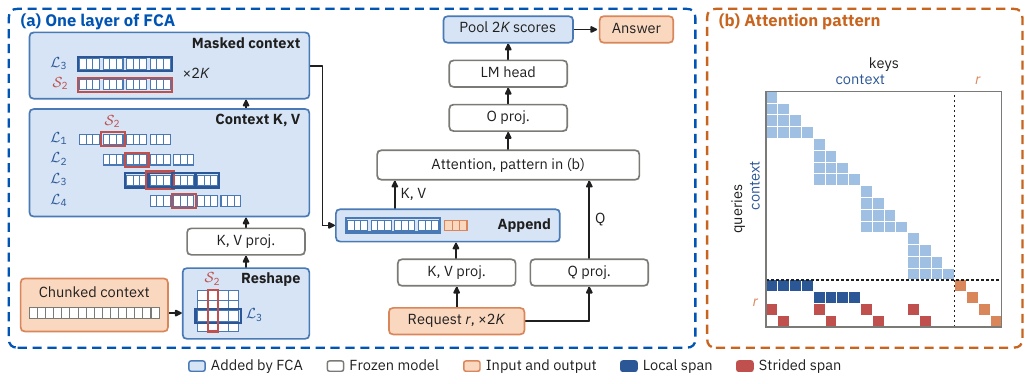}
  \caption{\textbf{One layer of \method.} (a) The chunked context is reshaped into a $K{\times}K$ grid, whose rows are local spans and whose columns are strided spans, and its keys and values are stored once. A request reads each span as a masked context, in $2K$ sequences that run in one batch and whose scores are pooled into the answer. (b) The resulting attention pattern for $K{=}4$.}
  \label{fig:fca}
\end{figure*}

\subsection{Native attention: Factorized Context Attention}
\label{sec:pattern}

\method computes the same $2K$ probes inside the model (\cref{fig:fca}). It splits the work of grid probing into two steps with two attention sets. The first step encodes the context once, and the second step answers each request from that encoding. Let $\Pre$ be the tokens of $p$, and identify a span with the set of its tokens.

\emph{Step 1, encoding the context once.} Every block lies in one local span, its row of the grid, and in one strided span, its column. \method encodes each block together with the earlier blocks of its row. Each prompt token attends causally within $\Pre$, and each token $t$ of a block in $\Loc_i$ attends to the prompt and to the earlier tokens of its own local span,
\begin{equation}
\mathcal{A}_{\mathrm{enc}}(t)=\Pre\cup\{t'\in\Loc_i : t'\le t\}.
\label{eq:enc}
\end{equation}
The $K$ rows are encoded in one batched pass, and the keys and values of every layer are stored. This store is the only encoding of the context.

\emph{Step 2, answering a request.} A request is answered by $2K$ short sequences, one for each span $T\in\{\Loc_1,\dots,\Loc_K,\Str_1,\dots,\Str_K\}$, which run in one batch. The sequence for $T$ gathers the stored keys and values of the blocks of $T$ and appends a copy $r_T$ of the request. Each token $u$ of $r_T$ attends to the prompt, to the gathered blocks of $T$, and to the earlier tokens of $r_T$,
\begin{equation}
\mathcal{A}_{T}(u)=\Pre\cup T\cup\{u'\in r_T : u'\le u\}.
\label{eq:ask}
\end{equation}
In a strided sequence, the request attends to all blocks of the span together, but each block was encoded with the earlier blocks of its row, not with the other blocks of its column. This is the one approximation of \method, and \cref{sec:ablations} measures it. The model runs only over the request tokens ($2K$ copies of $|r|$ tokens, one per span) and reads the score $\ell(\cdot\mid T)$ at the last token of $r_T$, as grid probing does. The stored keys and values of the spans are read and never changed, and the keys and values of the request copies are discarded once the scores are read. The next request is appended to the same store in the same way, so the context is encoded once for all requests.

\subsection{Stair positions}
\label{sec:stair}

In emulated attention every prompt places its span directly after $p$, so every span reads as contiguous text at small positions. In native attention the positions are fixed once, at encoding, and every sequence sees a block at the position it was encoded at. With rotary position encodings~\cite{rope}, attention depends on the distance between positions, so the positions must make every local and every strided span read as contiguous text. Two simple assignments fail. Placing every block at the start of the context gives all blocks of a span the same positions, so the model loses their order. Placing every block at its position in the document keeps each local span contiguous, but a strided span then jumps across the whole document, and the late spans of a long context fall beyond the trained window. Stair positions place the first token of the block at cell $(i,j)$ at
\begin{equation}
\pi(i,j)=|p|+(i+j-2)\,\sigma,
\label{eq:stair}
\end{equation}
where $\sigma$ is the length in tokens of the longest block, and the tokens of the block follow consecutively. A step along a row, from $(i,j)$ to $(i,j{+}1)$, and a step along a column, from $(i,j)$ to $(i{+}1,j)$, both add $\sigma$, so every row and every column reads as contiguous text, and each block sits at the same position in its row and in its column. The copy $r_T$ starts after the last block of its span. The largest position is about $2K\sigma\approx2\sqrt{sc}$, so a novel of 750k tokens in blocks of 500 tokens needs about 39k positions, inside the window of both text models we use. In a window of $W$ positions the stair positions therefore fit a context of up to $W^{2}/(4c)$ tokens, where the monolithic read fits $W$.

\begin{figure*}[!t]
  \centering
  \includegraphics[width=\textwidth]{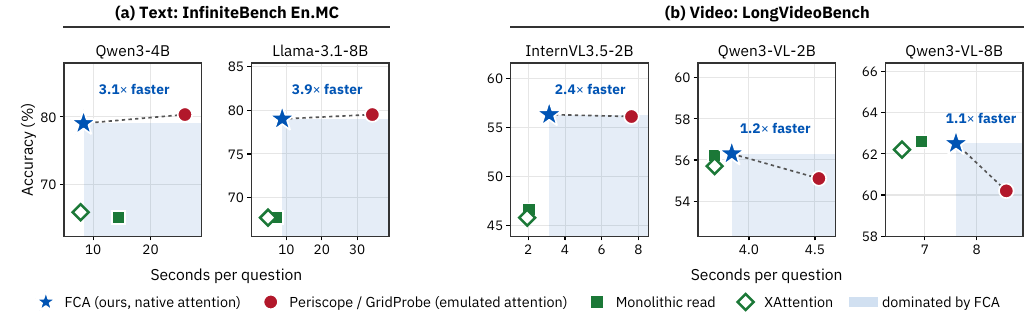}
  \caption{\textbf{Accuracy against time, one panel per model.} Time is the mean per question, where a read that reuses its encoding of a context (\method, the monolithic and XAttention) pays for that encoding once. The dashed link gives the speed-up of \method over the emulated read.}
  \label{fig:pareto}
\end{figure*}

\subsection{Properties and trade-offs}
\label{sec:exec}
\label{sec:emunative}

\begin{table}[t]
  \centering
  \footnotesize
  \renewcommand{\arraystretch}{1.0}
  \begin{tabular*}{\linewidth}{@{\extracolsep{\fill}}lll@{}}
    \toprule
    \textbf{Property} & \textbf{Emulated} & \textbf{Native} \\
    \midrule
    Accuracy, reach, map & same probes & same probes$^{*}$ \\
    Time, first question & slower or similar & \textbf{similar or faster} \\
    Time, later questions & as the first & \textbf{reuses encoding} \\
    New input & encodes all again & \textbf{new rows only} \\
    Memory & \textbf{one span} & the whole input \\
    Model access & \textbf{outputs only} & attention layers \\
    \bottomrule
  \end{tabular*}
  \caption{\textbf{Emulated attention (grid probing) against native attention (\method).} Bold marks the better side where the two differ. $^{*}$The strided spans of \method are approximate (\cref{sec:pattern}).}
  \label{tab:emunative}
\end{table}

Every sequence of \method is an ordinary sequence over a KV cache with a padding mask, so \method runs on standard attention kernels~\cite{flashattention2}. Emulated and native attention compute the same probes and differ in how they pay for them (\cref{tab:emunative}). Native attention gains two things from encoding the context once, in rows that do not depend on each other. First, the encoding is reused by every request, so its cost, which grows as $s^{1.5}$ as in grid probing~\cite{periscope}, is paid once. Second, new input becomes new rows of the grid, which are encoded at a cost that does not depend on what is already stored. Native attention pays for these gains in two ways. It keeps the encoding of the whole context in memory, as the monolithic read does, where emulated attention holds one span, and it needs the attention layers of the model, where emulated attention needs only its outputs. Emulated attention therefore suits closed models, tight memory budgets and single questions over inputs too large to keep, and native attention suits open models, inputs near or beyond the window, and repeated questions about one input.

Both kinds of attention share one property that the monolithic read lacks. Their rows and their span sequences do not depend on each other, so the work and, for native attention, the store can be divided among devices, unlike a causal pass over the whole context. Theoretically, with enough devices, the limit that remains is the position range of the model, which bounds the context at $W^{2}/(4c)$ tokens.

\section{Experiments}
\label{sec:experiments}

\begin{table*}[t]
  \centering
  \footnotesize
  \setlength{\tabcolsep}{2.6pt}
  \begin{tabular}{@{}l cccccc cccccc cccccc@{}}
    \toprule
    & \multicolumn{6}{c}{\textbf{InternVL3.5-2B}} & \multicolumn{6}{c}{\textbf{Qwen3-VL-2B}} & \multicolumn{6}{c}{\textbf{Qwen3-VL-8B}} \\
    \cmidrule(lr){2-7}\cmidrule(lr){8-13}\cmidrule(l){14-19}
    & \multicolumn{2}{c}{\textbf{Acc.} (\%) $\uparrow$} & \multicolumn{2}{c}{\textbf{TFLOPs} $\downarrow$} & \multicolumn{2}{c}{\textbf{TTFT} (s) $\downarrow$} & \multicolumn{2}{c}{\textbf{Acc.} (\%) $\uparrow$} & \multicolumn{2}{c}{\textbf{TFLOPs} $\downarrow$} & \multicolumn{2}{c}{\textbf{TTFT} (s) $\downarrow$} & \multicolumn{2}{c}{\textbf{Acc.} (\%) $\uparrow$} & \multicolumn{2}{c}{\textbf{TFLOPs} $\downarrow$} & \multicolumn{2}{c}{\textbf{TTFT} (s) $\downarrow$} \\
    \cmidrule(lr){2-3}\cmidrule(lr){4-5}\cmidrule(lr){6-7}\cmidrule(lr){8-9}\cmidrule(lr){10-11}\cmidrule(lr){12-13}\cmidrule(lr){14-15}\cmidrule(lr){16-17}\cmidrule(l){18-19}
    \textbf{Method} & LVB & V-MME & First & Later & Cold & Warm & LVB & V-MME & First & Later & Cold & Warm & LVB & V-MME & First & Later & Cold & Warm \\
    \midrule
    Monolithic & 46.4 & 44.8 & 463 & 463 & 3.67 & 3.67 & \underline{56.2} & \underline{47.1} & 868 & 868 & 6.40 & 6.40 & \textbf{62.7} & \underline{58.7} & 2574 & 2574 & 12.45 & 12.45 \\
    Monolithic, cached & 46.6 & 44.8 & 463 & 9 & 3.83 & 0.09 & \underline{56.2} & \textbf{47.3} & 868 & 12.1 & 7.06 & 0.14 & \underline{62.6} & \textbf{58.9} & 2574 & 36 & 13.11 & 0.26 \\
    XAttention~\cite{xattention}, cached & 45.8 & 42.8 & 247 & 9 & 3.65 & 0.09 & 55.7 & 46.4 & 486 & 12.1 & 7.07 & 0.13 & 62.2 & 58.1 & 1663 & 36 & 12.37 & 0.27 \\
    GridProbe~\cite{gridprobe} & \underline{56.1} & \textbf{46.8} & 521 & 521 & 7.62 & 7.62 & 55.1 & 46.7 & 298 & 298 & 4.53 & 4.53 & 60.2 & 55.8 & 1070 & 1070 & 8.57 & 8.57 \\
    \midrule
    \oursrow \textbf{\method}, direct & 50.8 & 44.8 & 229 & 80 & 3.43 & 0.62 & 51.7 & 44.8 & 191 & 76 & 2.87 & 0.59 & 55.8 & 51.8 & 741 & 359 & 5.88 & 1.66 \\
    \oursrow \textbf{\method}, select & \textbf{56.3} & \underline{45.8} & 341 & 192 & 4.45 & 1.72 & \textbf{56.3} & \textbf{47.3} & 378 & 262 & 5.05 & 2.60 & 62.5 & 56.2 & 1358 & 977 & 9.35 & 5.69 \\
    \bottomrule
  \end{tabular}
  \caption{\textbf{Long-video QA.} Accuracy (\%) on LongVideoBench (LVB) and Video-MME long (V-MME), 144 frames, and cost per LVB question on one A100: language-model TFLOPs by GridProbe's count and time to first token, for the first and a later question on a video. Cached reads keep a prefix cache. Bold marks the best and underline the second-best accuracy per model.}
  \label{tab:video}
\end{table*}

\subsection{Setup}
\label{sec:setup}

We compare three reads of the same frozen model, with the same prompt and read-out, the monolithic read as the reference, grid probing as the emulated read, and \method as the native read. The emulated read is Periscope for text~\cite{periscope}, GridProbe for video~\cite{gridprobe}, and the strips of AnswerMap for single images~\cite{answermap}. XAttention~\cite{xattention} and Token Sparse Attention~\cite{tokensparse}, two training-free sparse prefills of the monolithic read, serve as further references. Text experiments use Qwen3-4B~\cite{qwen3} and Llama-3.1-8B~\cite{llama3} on InfiniteBench En.MC~\cite{infinitebench}, 229 questions over novels with a median length of 150k tokens, and on SCBench~\cite{scbench}, with blocks of $c{=}500$ tokens. Video experiments use InternVL3.5-2B~\cite{wang2025internvl3}, Qwen3-VL-2B and 8B~\cite{bai2025qwen3} on LongVideoBench~\cite{wu2024longvideobench} and the long split of Video-MME~\cite{videomme}, with 144 uniformly sampled frames on a $12{\times}12$ grid, as in GridProbe. Image experiments follow AnswerMap with Qwen3-VL-4B on RefCOCOg~\cite{refcocog} and TextVQA~\cite{textvqa}. We report accuracy, the time to first token for the first question on a context (cold) and for later ones (warm), and peak GPU memory, all on one A100 80GB with FlashAttention-2~\cite{flashattention2}. We compare with grid probing directly, since its papers already compare it with the other training-free approaches, frame selection, agentic and streaming methods on video~\cite{gridprobe} and retrieval, agent, parallel-encoding and context-memory methods on text~\cite{periscope}, and a read that matches grid probing inherits these comparisons. Appendix gives the full setup.

\subsection{Accuracy and cost}
\label{sec:accuracy}
\label{sec:video}
\label{sec:text}

This section tests whether \method keeps the accuracy of grid probing at a lower cost. \Cref{fig:pareto} shows the trade-off behind all results. Grid probing buys its accuracy with time, and the monolithic read buys its speed with accuracy once the context exceeds the range that the model reads well. \method matches grid probing on accuracy at a fraction of its time.

On video (\cref{tab:video}), the gain of \method follows how much the context strains the model. InternVL3.5-2B writes every frame as 256 tokens, so 144 frames nearly fill its window and its monolithic read loses accuracy. There \method recovers the full accuracy of GridProbe, about 10 points above the monolithic read on LongVideoBench, in less than half the time of GridProbe. Qwen3-VL reads frames near their native resolution, well inside its window, so its monolithic read stays accurate. \method matches it on LongVideoBench, where GridProbe trails by 1 to 2 points, and only on Video-MME with Qwen3-VL-8B do both probing reads fall about 3 points behind, while both stay faster than the monolithic read on a first question. XAttention keeps the accuracy of the monolithic read but saves little time, because the vision encoder dominates the cost of a video read.

Long text shows the same pattern (\cref{tab:text}). The select read-out of \method stays within about a point of Periscope and 9 to 14 points above the monolithic read, and the one larger gap, in the direct read-out of Qwen3-4B, comes from the strided spans (\cref{sec:ablations}). The gain appears even on novels that fit the window, because every sequence of \method attends to only about $\sqrt{sc}$ tokens, the range in which these models are accurate~\cite{ruler,lostmiddle}. It widens on the novels beyond the window, where the monolithic read sees only a prefix and falls to 45.5 and 64.0 while \method stays at 70.5 and 80.0. XAttention and Token Sparse Attention add speed but not accuracy.

The speed of \method comes from encoding each novel once. Its first request is already faster than the monolithic prefill, since its encoding attends within spans, and every later request takes a tenth to a seventh of the time of Periscope. A cached monolithic read answers a later request faster still, at its lower accuracy, and \method generates tokens at the rate of the monolithic read.

\begin{table}[!t]
  \centering
  \footnotesize
  \setlength{\tabcolsep}{2.4pt}
  \begin{tabular}{@{}l cc cc c c@{}}
    \toprule
    & \multicolumn{2}{c}{\textbf{Accuracy} $\uparrow$} & \multicolumn{2}{c}{\textbf{TTFT} (s) $\downarrow$} & \textbf{TPOT} & \textbf{Mem.} \\
    \cmidrule(lr){2-3}\cmidrule(lr){4-5}
    \textbf{Method} & In-win. & All & Cold & Warm & (ms) & (GB) \\
    \midrule
    \multicolumn{7}{@{}l}{\emph{Qwen3-4B, window 262k}} \\
    Monolithic & 70.3 & 65.5 & 42.2 & 42.2 & 44.0 & 49.2 \\
    Monolithic, cached & 70.3 & 65.1 & 41.9 & 0.4 & 44.2 & 49.3 \\
    XAttention~\cite{xattention}, cached & 71.4 & 65.9 & 23.7 & 0.4 & 44.1 & 67.4 \\
    Token Sparse Attn.~\cite{tokensparse} & 68.1 & 63.8 & 37.9 & 37.9 & 45.4 & 67.3 \\
    Periscope~\cite{periscope}, direct & 80.0 & \underline{79.5} & 18.8 & 18.8 & -- & 11.7 \\
    Periscope~\cite{periscope}, select & \textbf{81.6} & \textbf{80.3} & 19.3 & 19.3 & 44.1 & 11.7 \\
    \cmidrule(l){1-7}
    \oursrow \textbf{\method}, direct & 77.3 & 75.1 & 14.3 & 2.3 & -- & 42.6 \\
    \oursrow \textbf{\method}, select & \underline{81.1} & 79.0 & 14.8 & 2.9 & 43.8 & 42.6 \\
    \midrule
    \multicolumn{7}{@{}l}{\emph{Llama-3.1-8B, window 131k}} \\
    Monolithic & 74.7 & 67.7 & 19.8 & 19.8 & 33.7 & 36.8 \\
    Monolithic, cached & 74.7 & 67.7 & 19.6 & 0.2 & 33.3 & 36.9 \\
    XAttention~\cite{xattention}, cached & 74.7 & 67.7 & 13.7 & 0.2 & 31.8 & 48.2 \\
    Token Sparse Attn.~\cite{tokensparse} & 74.7 & 68.1 & 20.5 & 20.5 & 33.0 & 48.1 \\
    Periscope~\cite{periscope}, direct & \textbf{81.0} & \textbf{80.8} & 16.6 & 16.6 & -- & 19.9 \\
    Periscope~\cite{periscope}, select & \underline{77.2} & 79.5 & 17.2 & 17.2 & 32.2 & 19.9 \\
    \cmidrule(l){1-7}
    \oursrow \textbf{\method}, direct & \textbf{81.0} & \underline{80.3} & 9.8 & 1.1 & -- & 41.8 \\
    \oursrow \textbf{\method}, select & \underline{77.2} & 79.0 & 10.4 & 1.7 & 31.4 & 41.8 \\
    \bottomrule
  \end{tabular}
  \caption{\textbf{Long-text QA on InfiniteBench En.MC.} Accuracy (\%) on all 229 questions and inside the window, and cost per request inside the window, on one A100. Bold marks the best and underline the second-best accuracy per column.}
  \label{tab:text}
\end{table}

\subsection{Reach beyond the window}
\label{sec:reach}

This section tests whether \method reads inputs that the monolithic read cannot hold.

\begin{table}[t]
  \centering
  \footnotesize
  \setlength{\tabcolsep}{2.1pt}
  \begin{tabular}{@{}l ccccc c@{}}
    \toprule
    & \multicolumn{5}{c}{\textbf{Accuracy} (\%) by frames $\uparrow$} & \textbf{Time} (s) \\
    \cmidrule(lr){2-6}\cmidrule(l){7-7}
    \textbf{Method} & 144 & 256 & 576 & 1{,}024 & 2{,}025 & 1{,}024 \\
    \midrule
    \multicolumn{7}{@{}l}{\emph{InternVL3.5-2B, window 40{,}960}} \\
    Monolithic, 144 frames$^{*}$ & 75.9 & 69.0 & 65.5 & 62.1 & 58.6 & 3.4 \\
    GridProbe~\cite{gridprobe} & \textbf{93.1} & \textbf{89.7} & \textbf{93.1} & \textbf{82.8} & \textbf{79.3} & 43.2 \\
    \oursrow \textbf{\method} (ours) & \textbf{93.1} & 86.2 & 86.2 & \textbf{82.8} & 75.9 & 26.0 \\
    \midrule
    \multicolumn{7}{@{}l}{\emph{Qwen3-VL-2B}} \\
    Monolithic read & 93.9 & 90.9 & 90.9 & 87.9 & 90.9 & 22.2 \\
    GridProbe~\cite{gridprobe} & \textbf{100.0} & \textbf{100.0} & 97.0 & \textbf{93.9} & \textbf{97.0} & 22.5 \\
    \oursrow \textbf{\method} (ours) & 97.0 & \textbf{100.0} & \textbf{100.0} & \textbf{93.9} & \textbf{97.0} & 21.6 \\
    \bottomrule
  \end{tabular}
  \caption{\textbf{Needle in a long video.} Accuracy (\%) on a question about one image placed as a frame of a Video-MME long video, and seconds per question at 1{,}024 frames, with the select read-out. Bold marks the best per model. $^{*}$Reads 144 frames at every length, the most its window holds.}
  \label{tab:vneedle}
\end{table}

\begin{figure*}[!t]
  \centering
  \includegraphics[width=\textwidth]{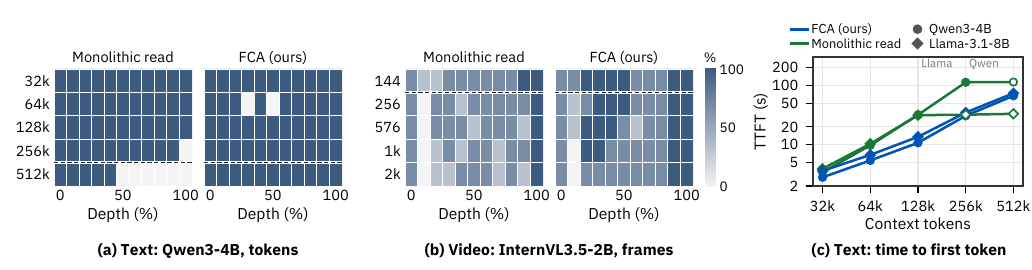}
  \caption{\textbf{Needles by length and depth.} Accuracy (\%) of the monolithic read and of \method, three needles per cell, with a dashed line at the window. (a) Text needles with Qwen3-4B, by context tokens. (b) Video needles with InternVL3.5-2B, by frames, over all needles. (c) Median time to first token of the first request on a text, with dotted lines at the windows. Hollow markers read only the window.}
  \label{fig:needle}
\end{figure*}

On video, an image from TextVQA~\cite{textvqa} replaces one frame of a Video-MME long video, and the model answers a question about that image (\cref{tab:vneedle,fig:needle}). InternVL3.5-2B fits at most 158 frames in its window, so its monolithic read samples 144 frames uniformly at every length, and these frames contain the needle less and less often. Its accuracy falls by about 17 points from 144 to 2{,}025 frames, while \method stays within one or two needles of GridProbe at about 60\% of its time. Qwen3-VL-2B shrinks its frames to fit, so its monolithic read degrades more gently and stays below \method.

On text, a sentence that pairs a word with a seven-digit number is inserted into the En.MC novels, cut to 32k to 512k tokens, following RULER~\cite{ruler} (\cref{tab:needle,fig:needle}). The monolithic read loses every needle beyond the prefix it keeps, while \method finds nearly all of them in contexts up to four times the window, and answers two to three times faster where both reads hold the whole context. Inside the window both reads find almost every needle, with the monolithic read slightly ahead.

\begin{table}[!h]
  \centering
  \small
  \setlength{\tabcolsep}{3.5pt}
  \begin{tabular}{@{}lccccc@{}}
    \toprule
    & \multicolumn{2}{c}{\textbf{Match} (\%) $\uparrow$} & \multicolumn{3}{c}{\textbf{TTFT} (s) $\downarrow$} \\
    \cmidrule(lr){2-3}\cmidrule(l){4-6}
    \textbf{Method} & In-window & Beyond & 128k & 256k & 512k \\
    \midrule
    \multicolumn{6}{@{}l}{\emph{Qwen3-4B, window 262k}} \\
    Monolithic read & \textbf{97.7} & 45.5 & 31.7 & 112.4 & 113.5$^{*}$ \\
    \oursrow \textbf{\method} (ours) & 95.5 & \textbf{100.0} & 10.7 & 30.6 & 66.4 \\
    \midrule
    \multicolumn{6}{@{}l}{\emph{Llama-3.1-8B, window 131k}} \\
    Monolithic read & \textbf{97.0} & 36.4 & 31.3 & 31.9$^{*}$ & 33.0$^{*}$ \\
    \oursrow \textbf{\method} (ours) & 93.9 & \textbf{98.5} & 13.5 & 34.7 & 73.4 \\
    \bottomrule
  \end{tabular}
  \caption{\textbf{Needle in a haystack.} String match (\%) of 165 needles in contexts of 32k to 512k tokens, and median time to first token of the first request on a context, with the select read-out of \method. Bold marks the higher accuracy per model. $^{*}$Reads only the prefix that fits the window.}
  \label{tab:needle}
\end{table}

\begin{table}[t]
  \centering
  \footnotesize
  \setlength{\tabcolsep}{2.4pt}
  \begin{tabular}{@{}l ccc ccc@{}}
    \toprule
    & \multicolumn{3}{c}{\textbf{Qwen3-4B}} & \multicolumn{3}{c}{\textbf{Llama-3.1-8B}} \\
    \cmidrule(lr){2-4}\cmidrule(l){5-7}
    \textbf{Method} & QA & MC & Time & QA & MC & Time \\
    \midrule
    Monolithic, prefix cache & 22.2 & 69.9 & 1.40 & 32.2 & 74.2 & 0.94 \\
    XAttention~\cite{xattention}, prefix cache & 19.2 & 69.9 & 1.56 & 32.1 & 74.2 & 0.93 \\
    Periscope~\cite{periscope}, select & 31.1 & 79.0 & 28.08 & 36.5 & 81.2 & 36.86 \\
    \midrule
    \oursrow \textbf{\method} (ours), select & 31.5 & 76.4 & 5.69 & 35.6 & 79.9 & 5.12 \\
    \bottomrule
  \end{tabular}
  \caption{\textbf{Multi-turn conversations on SCBench.} Token F1 (\%) on En.QA, accuracy (\%) on En.MC, over all turns, and seconds per later En.QA turn to the full answer.}
  \label{tab:multiturn}
\end{table}

\addtocounter{table}{1}
\begin{table*}[!t]
  \centering
  \small
  \setlength{\tabcolsep}{4pt}
  \begin{tabular}{@{}lc cccc c c cc@{}}
    \toprule
    & & \multicolumn{4}{c}{\textbf{Agreement}, RefCOCOg} & \textbf{Pointing} & \textbf{Deletion} & \multicolumn{2}{c}{\textbf{Time} (s)} \\
    \cmidrule(lr){3-6}\cmidrule(lr){9-10}
    \textbf{Method} & \textbf{Encodes} & NSS $\uparrow$ & AUC $\uparrow$ & Dist. $\downarrow$ & $r$ $\uparrow$ & In box (\%) $\uparrow$ & Changed (\%) $\uparrow$ & First & Later \\
    \midrule
    Random map & -- & $-$0.05 & 0.485 & 0.189 & 0.049 & 24.8 & 3.8 & -- & -- \\
    Image centre & -- & -- & -- & 0.189 & -- & 40.5 & -- & -- & -- \\
    \midrule
    AnswerMap strips~\cite{answermap}, $K{=}8$ & 16 & 1.32 & 0.826 & 0.130 & 0.664 & 68.5 & 53.4 & 0.289 & 0.289 \\
    AnswerMap strips, grids 3 and 5 & 16 & 1.54 & 0.853 & 0.122 & 0.711 & 71.0 & 53.1 & 0.384 & 0.384 \\
    \oursrow \textbf{\method} bands, $K{=}8$ & 1 & 1.38 & 0.777 & 0.118 & 0.731 & 71.5 & 42.5 & 0.217 & 0.143 \\
    \oursrow \textbf{\method} bands, grids 3 and 5 & 1 & 1.49 & 0.824 & 0.121 & 0.728 & 75.5 & 38.8 & 0.279 & 0.206 \\
    \bottomrule
  \end{tabular}
  \caption{\textbf{Single images in AnswerMap's protocol.} Qwen3-VL-4B with the image at 512 pixels. Agreement of the map with the point the model generates on 800 RefCOCOg expressions, map expectations inside the target box on 400, and TextVQA answers that change when the map's top 8 of 64 cells are blanked. Encodes counts image encodings per map. Time is seconds per map on 50 images and one A100, for the first question on an image and for a later question on it.}
  \label{tab:images}
\end{table*}

\begin{table*}[t]
  \centering
  \footnotesize
  \begin{subtable}[b]{0.3\linewidth}
    \centering
    \setlength{\tabcolsep}{3pt}
    \begin{tabular}{@{}lcc@{}}
      \toprule
      \textbf{Spans} & \textbf{Qwen} & \textbf{Llama} \\
      \midrule
      Local only & \textbf{6} & \textbf{14} \\
      Strided only & 46 & 40 \\
      \oursrow Both (default) & \underline{22} & \underline{21} \\
      \bottomrule
    \end{tabular}
    \caption{Answers changed vs.\ Periscope}
    \label{tab:abl-approx}
  \end{subtable}\hfill
  \begin{subtable}[b]{0.3\linewidth}
    \centering
    \setlength{\tabcolsep}{3pt}
    \begin{tabular}{@{}lcc@{}}
      \toprule
      \textbf{Encode} & \textbf{Qwen} & \textbf{Llama} \\
      \midrule
      Block alone & 68.6 & 63.8 \\
      \oursrow Local span (default) & 75.1 & \textbf{80.3} \\
      Local and strided & \textbf{78.2} & 79.5 \\
      \bottomrule
    \end{tabular}
    \caption{Encoding, direct acc.}
    \label{tab:abl-enc}
  \end{subtable}\hfill
  \begin{subtable}[b]{0.3\linewidth}
    \centering
    \setlength{\tabcolsep}{3pt}
    \begin{tabular}{@{}lcc@{}}
      \toprule
      \textbf{Positions} & \textbf{Qwen} & \textbf{Llama} \\
      \midrule
      Document & 72.1 & 48.0 \\
      \oursrow Stair (default) & \textbf{75.1} & \textbf{80.3} \\
      \bottomrule
    \end{tabular}
    \caption{Positions, direct acc.}
    \label{tab:abl-pos}
  \end{subtable}
  \caption{\textbf{Ablations on En.MC.} Each panel changes one part of \method, on all 229 questions, with accuracy in \%. Panel (a) counts the answers out of 229 that differ from Periscope. Qwen is Qwen3-4B and Llama is Llama-3.1-8B.}
  \label{tab:ablation}
\end{table*}
\addtocounter{table}{-3}

\begin{table}[!t]
  \centering
  \footnotesize
  \setlength{\tabcolsep}{3.2pt}
  \begin{tabular}{@{}l c cccc cc@{}}
    \toprule
    & & \multicolumn{4}{c}{\textbf{Match} (\%) by turn $\uparrow$} & \multicolumn{2}{c}{\textbf{Seconds} $\downarrow$} \\
    \cmidrule(lr){3-6}\cmidrule(l){7-8}
    \textbf{Method} & \textbf{All} & 1--4 & 5--8 & 9--12 & 13--16 & Encode & Turn \\
    \midrule
    \multicolumn{8}{@{}l}{\emph{Llama-3.1-8B, window 131k}} \\
    Monolithic & 53.1 & 93.8 & 50.0 & 37.5 & 31.2 & 18.7 & 19.8 \\
    \oursrow \textbf{\method} (ours) & \textbf{98.4} & \textbf{100.0} & \textbf{100.0} & \textbf{93.8} & \textbf{100.0} & \textbf{3.3} & \textbf{13.5} \\
    \midrule
    \multicolumn{8}{@{}l}{\emph{Qwen3-4B, window 262k}} \\
    Monolithic & 76.6 & 96.9 & 93.8 & 75.0 & 40.6 & 86.2 & 87.4 \\
    Periscope~\cite{periscope} & \textbf{100.0} & \textbf{100.0} & \textbf{100.0} & \textbf{100.0} & \textbf{100.0} & -- & 70.7 \\
    \oursrow \textbf{\method} (ours) & 95.3 & 93.8 & 96.9 & 90.6 & \textbf{100.0} & \textbf{2.7} & \textbf{13.6} \\
    \bottomrule
  \end{tabular}
  \caption{\textbf{Conversations with long turns.} 8 conversations of 16 turns, each turn adding 32k tokens, up to 524k tokens. String match (\%) of the select read-out over all turns and by turn. Encode is the seconds spent on encoding and Turn the seconds of the whole turn, both in turns 13 to 16. Periscope has no separate encoding.}
  \label{tab:append}
\end{table}

\addtocounter{table}{2}

\subsection{Many questions on one input}
\label{sec:openended}

This section tests whether the single encoding pays off when several questions are asked about one input, as on Video-MME with three questions per video and on LongVideoBench with about two.

On video, a later question reads the stored frames instead of encoding them again (\cref{tab:video}), and takes a quarter to two thirds of the time of GridProbe. On text, SCBench~\cite{scbench} asks up to five questions about one novel, open-ended or with four options (\cref{tab:multiturn}). \method keeps the accuracy of Periscope on open questions and stays within about three points of it on multiple choice, well above the monolithic read on both, and answers each later turn five to seven times faster than Periscope. A cached monolithic read remains the fastest per turn. When every turn adds new input, \method encodes only that input, at a cost and an accuracy that stay flat where the monolithic read has lost the early turns and Periscope is five times slower (\cref{tab:append}).

\subsection{Evidence localization}
\label{sec:images}

This section tests whether the map of \method finds the evidence. On single images, AnswerMap~\cite{answermap} asks a yes or no question about each row and column band of an image, each band shown as its own image, while \method asks the same questions over one encoding of the whole image (\cref{tab:images}). At $K{=}8$ the map of \method agrees with the point the model generates about as closely as the strips of AnswerMap (NSS 1.38 against 1.32) and lands in the target box for 71.5\% of the expressions against 68.5\%, from one encoding instead of 16 and in half the time for a later map. Blanking its top cells changes 42.5\% of the answers, against 53.4\% for the strips and 3.8\% for a random map, so the map of \method stays competitive with the strips and localizes the evidence at a fraction of their cost.

\section{Ablation Study}
\label{sec:ablations}

We ablate \method on En.MC, changing one part at a time and reporting the direct read-out, which is the output of the probes alone (\cref{tab:ablation}). \Cref{tab:abl-approx} locates the one approximation of \method. The local spans change only 6 and 14 of the 229 answers of Periscope, while the strided spans change 46 and 40, because a strided span reads blocks that were encoded within their local spans. \Cref{tab:abl-enc} shows how much the encoding of a block matters. Encoding a block within its local span instead of alone adds 6 to 16 points, and encoding the strided spans in their own pass as well removes the approximation and adds 3 points on Qwen3-4B, at twice the encoding cost. \Cref{tab:abl-pos} shows that stair positions are essential. With document positions the late spans fall outside the trained position range, and Llama-3.1-8B loses 32 points. The select read-out reads the chosen blocks again and recovers most of the loss of the strided spans (\cref{tab:text}).

\section{Conclusion and Limitations}
\label{sec:conclusion}

We showed that grid probing is emulated attention, an attention pattern reproduced from outside the model through prompts, and introduced \method, which runs the same pattern natively from a single encoding of the input. Across video, text and images, \method keeps the accuracy, the reach and the evidence maps of grid probing and answers every later request in a fraction of its time, with existing frozen models. \method has three limitations. It needs the attention layers of an open model, it keeps the encoding of the whole input in memory, and its strided probes approximate their emulated counterparts, so emulated attention remains the better choice for closed models and tight memory budgets. Future work can train a model with the pattern, run other prompting strategies natively, and emulate other training-free attention patterns through prompts. The independent rows also allow a store that grows with a stream, is divided among devices, or is queried as an index.

\section*{Acknowledgments}
We are grateful to the KAUST Academy for its generous support, and especially to Prof. Sultan Albarakati, who made this work possible. For computer time, this research used Ibex managed by the Supercomputing Core Laboratory at King Abdullah University of Science \& Technology (KAUST) in Thuwal, Saudi Arabia.

{
    \small
    \setlength{\bibsep}{0pt}
    \bibliographystyle{ieeenat_fullname}
    \bibliography{main}

\begin{thebibliography}{48}
\providecommand{\natexlab}[1]{#1}
\providecommand{\url}[1]{\texttt{#1}}
\expandafter\ifx\csname urlstyle\endcsname\relax
  \providecommand{\doi}[1]{doi: #1}\else
  \providecommand{\doi}{doi: \begingroup \urlstyle{rm}\Url}\fi

\bibitem[Acharya et~al.(2024)Acharya, Jia, and Ginsburg]{starattention}
Shantanu Acharya, Fei Jia, and Boris Ginsburg.
\newblock Star attention: Efficient {LLM} inference over long sequences.
\newblock \emph{arXiv preprint arXiv:2411.17116}, 2024.

\bibitem[Bai et~al.(2025{\natexlab{a}})Bai, Cai, Chen, Chen, Chen, Cheng, Deng,
  Ding, Gao, Ge, et~al.]{bai2025qwen3}
Shuai Bai, Yuxuan Cai, Ruizhe Chen, Keqin Chen, Xionghui Chen, Zesen Cheng,
  Lianghao Deng, Wei Ding, Chang Gao, Chunjiang Ge, et~al.
\newblock {Qwen3-VL} technical report.
\newblock \emph{arXiv preprint arXiv:2511.21631}, 2025{\natexlab{a}}.

\bibitem[Bai et~al.(2025{\natexlab{b}})Bai, Tu, Zhang, Peng, Wang, Lv, Cao, Xu,
  Hou, Dong, Tang, and Li]{longbenchv2}
Yushi Bai, Shangqing Tu, Jiajie Zhang, Hao Peng, Xiaozhi Wang, Xin Lv, Shulin
  Cao, Jiazheng Xu, Lei Hou, Yuxiao Dong, Jie Tang, and Juanzi Li.
\newblock {LongBench} v2: Towards deeper understanding and reasoning on
  realistic long-context multitasks.
\newblock In \emph{Proceedings of the 63rd Annual Meeting of the Association
  for Computational Linguistics (Volume 1: Long Papers)}, pages 3639--3664,
  2025{\natexlab{b}}.

\bibitem[Beltagy et~al.(2020)Beltagy, Peters, and Cohan]{longformer}
Iz Beltagy, Matthew~E. Peters, and Arman Cohan.
\newblock Longformer: The long-document transformer.
\newblock \emph{arXiv preprint arXiv:2004.05150}, 2020.

\bibitem[Child et~al.(2019)Child, Gray, Radford, and
  Sutskever]{sparsetransformer}
Rewon Child, Scott Gray, Alec Radford, and Ilya Sutskever.
\newblock Generating long sequences with sparse transformers.
\newblock \emph{arXiv preprint arXiv:1904.10509}, 2019.

\bibitem[Dao(2024)]{flashattention2}
Tri Dao.
\newblock {FlashAttention-2}: Faster attention with better parallelism and work
  partitioning.
\newblock In \emph{ICLR}, 2024.

\bibitem[Dao et~al.(2022)Dao, Fu, Ermon, Rudra, and R{\'e}]{flashattention}
Tri Dao, Daniel~Y. Fu, Stefano Ermon, Atri Rudra, and Christopher R{\'e}.
\newblock {FlashAttention}: Fast and memory-efficient exact attention with
  {IO}-awareness.
\newblock In \emph{NeurIPS}, 2022.

\bibitem[Dong et~al.(2024)Dong, Feng, Guessous, Liang, and He]{flexattention}
Juechu Dong, Boyuan Feng, Driss Guessous, Yanbo Liang, and Horace He.
\newblock Flex attention: A programming model for generating optimized
  attention kernels.
\newblock \emph{arXiv preprint arXiv:2412.05496}, 2024.

\bibitem[Du et~al.(2025)Du, Tian, Ronanki, Rongali, Bodapati, Galstyan, Wells,
  Schwartz, Huerta, and Peng]{ducontext}
Yufeng Du, Minyang Tian, Srikanth Ronanki, Subendhu Rongali, Sravan~Babu
  Bodapati, Aram Galstyan, Azton Wells, Roy Schwartz, Eliu~A Huerta, and Hao
  Peng.
\newblock Context length alone hurts {LLM} performance despite perfect
  retrieval.
\newblock In \emph{Findings of the Association for Computational Linguistics:
  EMNLP 2025}, pages 23281--23298, 2025.

\bibitem[Eltahir et~al.(2026{\natexlab{a}})Eltahir, Adam, Tahir, Alamoudi,
  Ammar, Alboloshi, Albluey, Hussain, and Khan]{answermap}
Mohamed Eltahir, Fardows Adam, Duaa~M. Tahir, Lama Alamoudi, Sana Ammar,
  Atheer~A. Alboloshi, Jory Albluey, Tanveer Hussain, and Naeemullah Khan.
\newblock {AnswerMap}: Faithful spatial interpretability of {VLMs} from answer
  posteriors.
\newblock \emph{arXiv preprint arXiv:2609.35247}, 2026{\natexlab{a}}.

\bibitem[Eltahir et~al.(2026{\natexlab{b}})Eltahir, Ayash, Habibullah, Hussain,
  and Khan]{gridprobe}
Mohamed Eltahir, Lama Ayash, Ali Habibullah, Tanveer Hussain, and Naeemullah
  Khan.
\newblock {GridProbe}: Posterior-probing for adaptive test-time compute in
  long-video {VLMs}.
\newblock In \emph{NeurIPS}, 2026{\natexlab{b}}.

\bibitem[Eltahir et~al.(2026{\natexlab{c}})Eltahir, Obayd, Rashid, Alghamdi,
  Mousa, Ahmed, Hussain, and Khan]{periscope}
Mohamed Eltahir, Anas Obayd, Raed Rashid, Abdulrahman Alghamdi, Abdulrahman
  Mousa, Abdallah Ahmed, Tanveer Hussain, and Naeemullah Khan.
\newblock Periscope: Extending frozen language models beyond their context
  window.
\newblock \emph{arXiv preprint arXiv:2610.04047}, 2026{\natexlab{c}}.

\bibitem[Fu et~al.(2025)Fu, Dai, Luo, Li, Ren, Zhang, Wang, Zhou, Shen, Zhang,
  et~al.]{videomme}
Chaoyou Fu, Yuhan Dai, Yongdong Luo, Lei Li, Shuhuai Ren, Renrui Zhang, Zihan
  Wang, Chenyu Zhou, Yunhang Shen, Mengdan Zhang, et~al.
\newblock {Video-MME}: The first-ever comprehensive evaluation benchmark of
  multi-modal {LLMs} in video analysis.
\newblock In \emph{CVPR}, 2025.

\bibitem[Grattafiori et~al.(2024)Grattafiori, Dubey, Jauhri, Pandey, Kadian,
  et~al.]{llama3}
Aaron Grattafiori, Abhimanyu Dubey, Abhinav Jauhri, Abhinav Pandey, Abhishek
  Kadian, et~al.
\newblock The {Llama} 3 herd of models.
\newblock \emph{arXiv preprint arXiv:2407.21783}, 2024.

\bibitem[Ho et~al.(2019)Ho, Kalchbrenner, Weissenborn, and Salimans]{axial}
Jonathan Ho, Nal Kalchbrenner, Dirk Weissenborn, and Tim Salimans.
\newblock Axial attention in multidimensional transformers.
\newblock \emph{arXiv preprint arXiv:1912.12180}, 2019.

\bibitem[Hsieh et~al.(2024)Hsieh, Sun, Kriman, Acharya, Rekesh, Jia, Zhang, and
  Ginsburg]{ruler}
Cheng-Ping Hsieh, Simeng Sun, Samuel Kriman, Shantanu Acharya, Dima Rekesh, Fei
  Jia, Yang Zhang, and Boris Ginsburg.
\newblock {RULER}: What's the real context size of your long-context language
  models?
\newblock \emph{arXiv preprint arXiv:2404.06654}, 2024.

\bibitem[Hu et~al.(2025)]{epic}
Junhao Hu et~al.
\newblock {EPIC}: Efficient position-independent caching for serving large
  language models.
\newblock In \emph{ICML}, 2025.

\bibitem[Jiang et~al.(2024)Jiang, Li, Zhang, Wu, Luo, et~al.]{minference}
Huiqiang Jiang, Yucheng Li, Chengruidong Zhang, Qianhui Wu, Xufang Luo, et~al.
\newblock {MInference} 1.0: Accelerating pre-filling for long-context {LLMs}
  via dynamic sparse attention.
\newblock In \emph{NeurIPS}, 2024.

\bibitem[Jo et~al.(2026)Jo, Kang, Song, and Kim]{tokensparse}
Dongwon Jo, Beomseok Kang, Jiwon Song, and Jae-Joon Kim.
\newblock Token sparse attention: Efficient long-context inference with
  interleaved token selection.
\newblock In \emph{Proceedings of the 43rd International Conference on Machine
  Learning}, 2026.

\bibitem[Juravsky et~al.(2024)Juravsky, Brown, Ehrlich, Fu, R{\'e}, and
  Mirhoseini]{hydragen}
Jordan Juravsky, Bradley Brown, Ryan Ehrlich, Daniel~Y. Fu, Christopher R{\'e},
  and Azalia Mirhoseini.
\newblock Hydragen: High-throughput {LLM} inference with shared prefixes.
\newblock \emph{arXiv preprint arXiv:2402.05099}, 2024.

\bibitem[Kwon et~al.(2023)Kwon, Li, Zhuang, Sheng, Zheng, Yu, Gonzalez, Zhang,
  and Stoica]{kwon2023pagedattention}
Woosuk Kwon, Zhuohan Li, Siyuan Zhuang, Ying Sheng, Lianmin Zheng, Cody~Hao Yu,
  Joseph~E. Gonzalez, Hao Zhang, and Ion Stoica.
\newblock Efficient memory management for large language model serving with
  {PagedAttention}.
\newblock In \emph{Proceedings of the 29th Symposium on Operating Systems
  Principles}, 2023.

\bibitem[Li et~al.(2025)Li, Jiang, Wu, Luo, Ahn, Zhang, Abdi, Li, Gao, Yang,
  and Qiu]{scbench}
Yucheng Li, Huiqiang Jiang, Qianhui Wu, Xufang Luo, Surin Ahn, Chengruidong
  Zhang, Amir~H. Abdi, Dongsheng Li, Jianfeng Gao, Yuqing Yang, and Lili Qiu.
\newblock {SCBench}: A {KV} cache-centric analysis of long-context methods.
\newblock In \emph{ICLR}, 2025.

\bibitem[Liu et~al.(2025{\natexlab{a}})Liu, Mei, Lin, Xue, Wang, Xu, Wu, Zhang,
  Lin, Dong, et~al.]{deepseekv32}
Aixin Liu, Aoxue Mei, Bangcai Lin, Bing Xue, Bingxuan Wang, Bingzheng Xu,
  Bochao Wu, Bowei Zhang, Chaofan Lin, Chen Dong, et~al.
\newblock {DeepSeek-V3.2}: Pushing the frontier of open large language models.
\newblock \emph{arXiv preprint arXiv:2512.02556}, 2025{\natexlab{a}}.

\bibitem[Liu et~al.(2024)Liu, Lin, Hewitt, Paranjape, Bevilacqua, Petroni, and
  Liang]{lostmiddle}
Nelson~F. Liu, Kevin Lin, John Hewitt, Ashwin Paranjape, Michele Bevilacqua,
  Fabio Petroni, and Percy Liang.
\newblock Lost in the middle: How language models use long contexts.
\newblock \emph{Transactions of the Association for Computational Linguistics},
  12:\penalty0 157--173, 2024.

\bibitem[Liu et~al.(2025{\natexlab{b}})]{vidcom2}
Xuyang Liu et~al.
\newblock Video compression commander: Plug-and-play inference acceleration for
  video large language models.
\newblock In \emph{EMNLP}, 2025{\natexlab{b}}.

\bibitem[Lu et~al.(2025)Lu, Jiang, Liu, Du, Jiang, et~al.]{moba}
Enzhe Lu, Zhejun Jiang, Jingyuan Liu, Yulun Du, Tao Jiang, et~al.
\newblock {MoBA}: Mixture of block attention for long-context {LLMs}.
\newblock In \emph{NeurIPS}, 2025.

\bibitem[Ma et~al.(2025)Ma, Wang, and Tian]{blockattention}
Dongyang Ma, Yan Wang, and Lan Tian.
\newblock Block-attention for efficient prefilling.
\newblock In \emph{ICLR}, 2025.

\bibitem[Mao et~al.(2016)Mao, Huang, Toshev, Camburu, Yuille, and
  Murphy]{refcocog}
Junhua Mao, Jonathan Huang, Alexander Toshev, Oana Camburu, Alan~L. Yuille, and
  Kevin Murphy.
\newblock Generation and comprehension of unambiguous object descriptions.
\newblock In \emph{CVPR}, 2016.

\bibitem[Modarressi et~al.(2025)Modarressi, Deilamsalehy, Dernoncourt, Bui,
  Rossi, Yoon, and Schuetze]{nolima}
Ali Modarressi, Hanieh Deilamsalehy, Franck Dernoncourt, Trung Bui, Ryan~A.
  Rossi, Seunghyun Yoon, and Hinrich Schuetze.
\newblock {NoLiMa}: Long-context evaluation beyond literal matching.
\newblock In \emph{ICML}, pages 44554--44570, 2025.

\bibitem[Peng et~al.(2024)Peng, Quesnelle, Fan, and Shippole]{yarn}
Bowen Peng, Jeffrey Quesnelle, Honglu Fan, and Enrico Shippole.
\newblock {YaRN}: Efficient context window extension of large language models.
\newblock In \emph{ICLR}, 2024.

\bibitem[Ratner et~al.(2023)Ratner, Levine, Belinkov, Ram, Magar, Abend,
  Karpas, Shashua, Leyton-Brown, and Shoham]{pcw}
Nir Ratner, Yoav Levine, Yonatan Belinkov, Ori Ram, Inbal Magar, Omri Abend,
  Ehud Karpas, Amnon Shashua, Kevin Leyton-Brown, and Yoav Shoham.
\newblock Parallel context windows for large language models.
\newblock In \emph{Proceedings of the 61st Annual Meeting of the Association
  for Computational Linguistics (Volume 1: Long Papers)}, pages 6383--6402,
  2023.

\bibitem[Shao et~al.(2025)]{holitom}
Kele Shao et~al.
\newblock {HoliTom}: Holistic token merging for fast video large language
  models.
\newblock In \emph{NeurIPS}, 2025.

\bibitem[Shen et~al.(2025)]{fastvid}
Leqi Shen et~al.
\newblock {FastVID}: Dynamic density pruning for fast video large language
  models.
\newblock In \emph{NeurIPS}, 2025.

\bibitem[Singh et~al.(2019)Singh, Natarajan, Shah, Jiang, Chen, Batra, Parikh,
  and Rohrbach]{textvqa}
Amanpreet Singh, Vivek Natarajan, Meet Shah, Yu Jiang, Xinlei Chen, Dhruv
  Batra, Devi Parikh, and Marcus Rohrbach.
\newblock Towards {VQA} models that can read.
\newblock In \emph{CVPR}, 2019.

\bibitem[Su et~al.(2024)Su, Ahmed, Lu, Pan, Bo, and Liu]{rope}
Jianlin Su, Murtadha Ahmed, Yu Lu, Shengfeng Pan, Wen Bo, and Yunfeng Liu.
\newblock {RoFormer}: Enhanced transformer with rotary position embedding.
\newblock \emph{Neurocomputing}, 568:\penalty0 127063, 2024.

\bibitem[Wang et~al.(2025)Wang, Gao, Gu, Pu, Cui, Wei, Liu, Jing, Ye, Shao,
  et~al.]{wang2025internvl3}
Weiyun Wang, Zhangwei Gao, Lixin Gu, Hengjun Pu, Long Cui, Xingguang Wei,
  Zhaoyang Liu, Linglin Jing, Shenglong Ye, Jie Shao, et~al.
\newblock {InternVL3.5}: Advancing open-source multimodal models in
  versatility, reasoning, and efficiency.
\newblock \emph{arXiv preprint arXiv:2508.18265}, 2025.

\bibitem[Wolf et~al.(2020)Wolf, Debut, Sanh, Chaumond, Delangue, Moi, Cistac,
  Rault, Louf, Funtowicz, et~al.]{transformers}
Thomas Wolf, Lysandre Debut, Victor Sanh, Julien Chaumond, Clement Delangue,
  Anthony Moi, Pierric Cistac, Tim Rault, R{\'e}mi Louf, Morgan Funtowicz,
  et~al.
\newblock Transformers: State-of-the-art natural language processing.
\newblock In \emph{Proceedings of the 2020 Conference on Empirical Methods in
  Natural Language Processing: System Demonstrations}, pages 38--45, 2020.

\bibitem[Wu et~al.(2024)Wu, Li, Chen, and Li]{wu2024longvideobench}
Haoning Wu, Dongxu Li, Bei Chen, and Junnan Li.
\newblock {LongVideoBench}: A benchmark for long-context interleaved
  video-language understanding.
\newblock In \emph{NeurIPS}, pages 28828--28857, 2024.

\bibitem[Xiao et~al.(2024{\natexlab{a}})Xiao, Zhang, Han, Xiao, Lin, Zhang,
  Liu, and Sun]{infllm}
Chaojun Xiao, Pengle Zhang, Xu Han, Guangxuan Xiao, Yankai Lin, Zhengyan Zhang,
  Zhiyuan Liu, and Maosong Sun.
\newblock {InfLLM}: Training-free long-context extrapolation for {LLMs} with an
  efficient context memory.
\newblock In \emph{NeurIPS}, pages 119638--119661, 2024{\natexlab{a}}.

\bibitem[Xiao et~al.(2024{\natexlab{b}})Xiao, Tian, Chen, Han, and
  Lewis]{streamingllm}
Guangxuan Xiao, Yuandong Tian, Beidi Chen, Song Han, and Mike Lewis.
\newblock Efficient streaming language models with attention sinks.
\newblock In \emph{ICLR}, 2024{\natexlab{b}}.

\bibitem[Xu et~al.(2025)Xu, Xiao, Huang, Guo, and Han]{xattention}
Ruyi Xu, Guangxuan Xiao, Haofeng Huang, Junxian Guo, and Song Han.
\newblock {XAttention}: Block sparse attention with antidiagonal scoring.
\newblock In \emph{Proceedings of the 42nd International Conference on Machine
  Learning}, pages 69819--69831, 2025.

\bibitem[Yang et~al.(2025{\natexlab{a}})Yang, Li, Yang, Zhang, Hui, Zheng, Yu,
  Gao, Huang, Lv, et~al.]{qwen3}
An Yang, Anfeng Li, Baosong Yang, Beichen Zhang, Binyuan Hui, Bo Zheng, Bowen
  Yu, Chang Gao, Chengen Huang, Chenxu Lv, et~al.
\newblock {Qwen3} technical report.
\newblock \emph{arXiv preprint arXiv:2505.09388}, 2025{\natexlab{a}}.

\bibitem[Yang et~al.(2025{\natexlab{b}})]{visionzip}
Senqiao Yang et~al.
\newblock {VisionZip}: Longer is better but not necessary in vision language
  models.
\newblock In \emph{CVPR}, 2025{\natexlab{b}}.

\bibitem[Yang et~al.(2025{\natexlab{c}})Yang, Chen, and Chen]{ape}
Xinyu Yang, Tianqi Chen, and Beidi Chen.
\newblock {APE}: Faster and longer context-augmented generation via adaptive
  parallel encoding.
\newblock In \emph{ICLR}, 2025{\natexlab{c}}.

\bibitem[Yao et~al.(2025)Yao, Li, Liu, Ray, Cheng, Zhang, Du, Lu, and
  Jiang]{cacheblend}
Jiayi Yao, Hanchen Li, Yuhan Liu, Siddhant Ray, Yihua Cheng, Qizheng Zhang,
  Kuntai Du, Shan Lu, and Junchen Jiang.
\newblock {CacheBlend}: Fast large language model serving for {RAG} with cached
  knowledge fusion.
\newblock In \emph{Proceedings of the Twentieth European Conference on Computer
  Systems}, 2025.

\bibitem[Yuan et~al.(2025)Yuan, Gao, Dai, Luo, Zhao, et~al.]{nsa}
Jingyang Yuan, Huazuo Gao, Damai Dai, Junyu Luo, Liang Zhao, et~al.
\newblock Native sparse attention: Hardware-aligned and natively trainable
  sparse attention.
\newblock In \emph{Proceedings of the 63rd Annual Meeting of the Association
  for Computational Linguistics (Volume 1: Long Papers)}, 2025.

\bibitem[Zaheer et~al.(2020)Zaheer, Guruganesh, Dubey, Ainslie, Alberti,
  Ontanon, Pham, Ravula, Wang, Yang, and Ahmed]{bigbird}
Manzil Zaheer, Guru Guruganesh, Kumar~Avinava Dubey, Joshua Ainslie, Chris
  Alberti, Santiago Ontanon, Philip Pham, Anirudh Ravula, Qifan Wang, Li Yang,
  and Amr Ahmed.
\newblock Big bird: Transformers for longer sequences.
\newblock In \emph{NeurIPS}, 2020.

\bibitem[Zhang et~al.(2024)Zhang, Chen, Hu, Xu, Chen, Hao, Han, Thai, Wang,
  Liu, et~al.]{infinitebench}
Xinrong Zhang, Yingfa Chen, Shengding Hu, Zihang Xu, Junhao Chen, Moo Hao, Xu
  Han, Zhen Thai, Shuo Wang, Zhiyuan Liu, et~al.
\newblock $\infty${Bench}: Extending long context evaluation beyond 100{K}
  tokens.
\newblock In \emph{Proceedings of the 62nd Annual Meeting of the Association
  for Computational Linguistics (Volume 1: Long Papers)}, pages 15262--15277,
  2024.

\end{thebibliography}
}

\clearpage
\ifdefined\arxivversion\else\setcounter{page}{1}\fi
\maketitlesupplementary
\appendix

This supplementary material follows the order of the paper. \Cref{app:impl,app:cost} give the implementation and the cost model of \cref{sec:method}, and \cref{app:setup} gives the full setup of \cref{sec:experiments}. \Cref{app:accuracy,app:reach,app:many,app:local} then complete the four groups of results of \cref{sec:experiments} in the same setting, with the remaining splits of Video-MME, a larger InternVL model, token reduction next to the sparse prefill of \cref{tab:video}, and Periscope on the text needles. \Cref{app:abl} completes the ablations of \cref{sec:ablations}, and \cref{app:limits} states the scope of \method. Every result uses the models, frames, prompts and reads of \cref{sec:experiments}.

\section{Implementation}
\label{app:impl}

\paragraph{Prompt and read-out.} For text, a probe reads the chat template of the model around
\begin{quote}\small\ttfamily
Read the book and answer the question.\\[2pt]
\{span\}\\[2pt]
Question: \{question\}\\
A. \{option\} \dots\ D. \{option\}\\
E. Unsure\\
Answer with a single letter:
\end{quote}
with thinking disabled where the template has it. The probe score of \cref{eq:probe} is computed from the top-20 log-probabilities of the first answer token, as in Periscope~\cite{periscope}. The select read and the monolithic read use the same prompt without the abstain option. Blocks are chunks of 500 tokens of the model's own tokenizer, and $K{=}\lceil\sqrt{N}\,\rceil$.

\paragraph{Store.} The keys and values of all layers sit in one buffer of shape $(2, \text{layers}, |p|{+}s, \text{heads}, d)$, token-major, with the prompt in the first $|p|$ rows and each block at its offset. Encoding runs one causal pass per local span over the prompt and the span at stair positions, and writes the prompt once and every block once. The buffer stays on the GPU when it fits 24GB, and otherwise is allocated in host memory and page-locked, so every run of consecutive blocks is gathered in one direct transfer. The buffer is released when the context changes.

\paragraph{Batching.} The $2K$ sequences of a request are sorted by span length and run in batches of $B$. For each layer the gathered spans of a batch form one right-aligned cache with zeros in the padding slots, a two-dimensional mask hides the padding, and every row carries the positions of its own request copy. With $B{=}1$ the computation reduces to one sequence at a time.

\paragraph{Monolithic read with prefix cache.} The context is cut once per context to the window minus the prompt and a margin of 1{,}024 tokens, so the cut does not depend on the question. Its tokens are taken as the common prefix of two full prompts with different questions, so they equal the tokens of the monolithic read exactly. The cache is preallocated, filled in pieces of 32{,}768 tokens, and cut back to the context after every request.

\paragraph{Generation timing.} TPOT decodes 100 greedy tokens through the cache of each read after its answer token. The generated text is discarded and its time is excluded from TTFT.

\paragraph{Video.} Each frame is one block. Qwen3-VL fuses two consecutive frames into one temporal group. The store therefore repeats every frame over its own group, so the tokens of a frame do not depend on the neighbouring frame and every column of the grid stays distinct. The store holds Qwen3-VL frames at $448{\times}448$ pixels with the timestamps of the monolithic read, and InternVL3.5 frames at $224{\times}224$ as one tile of 256 tokens. Stair positions offset all three axes of the multi-axis rotary encoding of Qwen3-VL and the single axis of InternVL3.5. Every probe carries GridProbe's request, with the subtitles on LongVideoBench, and the select read-out is GridProbe's focused pass on the frames of the map's top cells, with GridProbe's rule for their number, at native resolution for Qwen3-VL and at $224{\times}224$ for InternVL3.5 as in GridProbe~\cite{gridprobe}. XAttention runs with stride 8 and threshold 0.9 on the attention of the language model only, since the vision encoder attends within each frame.

\section{Execution and cost model}
\label{app:cost}

\paragraph{Execution.} Every sequence of a request is an ordinary padded sequence over a key-value cache (\cref{app:impl}), so the pattern runs on standard attention kernels~\cite{flashattention2} without change. For generation, the select read-out reads the top $K$ blocks of $M$ again with the request and decodes over them, so each generated token attends to about $\sqrt{sc}$ keys rather than $s$.

\paragraph{Cost model.} To compare costs, we follow~\cite{periscope} and let a forward over $t$ tokens cost $at+bt^{2}$ FLOPs, where $a$ is the cost per token of the layers that act on one token at a time and $b$ is the attention cost per pair of tokens. \Cref{tab:cost} compares the three reads under this model. The encoding of \method runs $K$ local spans of about $\sqrt{sc}$ tokens each, so its attention grows as $\sqrt{c}\,s^{1.5}$ rather than $s^{2}$, the scaling that grid probing obtains from its short spans~\cite{periscope}, and it costs less than a monolithic prefill for every context longer than one block. A request then costs $2K$ sequences of $|r|$ tokens over about $\sqrt{sc}$ keys each, since every block is read once in its local span and once in its strided span. Grid probing pays for the encoding again on every request.

\begin{table}[h]
  \centering
  \footnotesize
  \setlength{\tabcolsep}{2.5pt}
  \begin{tabular}{@{}lccc@{}}
    \toprule
    \textbf{Read} & \textbf{Encode (once)} & \textbf{Per request} & \textbf{GPU cache} \\
    \midrule
    Monolithic, prefix cache & $as+bs^{2}$ & $a|r|+b|r|s$ & $s$ \\
    Grid probing, emulated & -- & $2as+2b\sqrt{c}\,s^{1.5}$ & $\sqrt{sc}$ \\
    \oursrow \textbf{\method}, native & $as+b\sqrt{c}\,s^{1.5}$ & $2Ka|r|+2b|r|s$ & $B\sqrt{sc}^{\,*}$ \\
    \bottomrule
  \end{tabular}
  \caption{\textbf{Cost of the three reads.} FLOPs and GPU cache for a context of $s$ tokens in blocks of $c$, a request of $|r|$ tokens, and $K{\approx}\sqrt{s/c}$, without the prompt and the select read, where a forward over $t$ tokens costs $at+bt^{2}$. $^{*}$The store of $s$ tokens sits in host memory when it does not fit the GPU.}
  \label{tab:cost}
\end{table}

\subsection{Cost and context length}
\label{app:costlength}

\begin{figure}[h]
  \centering
  \includegraphics[width=0.62\linewidth]{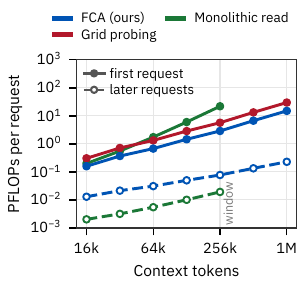}
  \caption{\textbf{Cost and context length.} FLOPs per request of Qwen3-4B against context length, for the first request on a context and a later one, from the cost model of \cref{tab:cost}.}
  \label{fig:costlength}
\end{figure}

\Cref{fig:costlength} evaluates the cost model of \cref{tab:cost} for Qwen3-4B. The monolithic prefill grows with the square of the context up to the window, while the encoding of \method grows as $s^{1.5}$ and a later request costs only $2K$ short sequences. The gap between the two reads therefore widens as the context grows.

\subsection{Attention kernels}
\label{app:kernels}

\method needs no special kernel. It runs unchanged on FlexAttention~\cite{flexattention}, and on the En.MC questions read on one node, Qwen3-4B gives the same accuracy under FlexAttention as under FlashAttention-2, within the noise floor of \cref{app:noise}, at somewhat more time per question.

\section{Experimental setup}
\label{app:setup}

\paragraph{Models and benchmarks.} Text experiments use Qwen3-4B-Instruct-2507~\cite{qwen3}, with a window of 262k tokens, and Llama-3.1-8B-Instruct~\cite{llama3}, with a window of 131k. The main benchmark is InfiniteBench En.MC~\cite{infinitebench}, 229 four-option questions over 58 novels with a median length of 150k tokens and a longest novel near 750k. SCBench~\cite{scbench} tests conversations over the same novels, and LongBench v2~\cite{longbenchv2} serves as a second benchmark (\cref{app:longbench}). Blocks are chunks of $c{=}500$ tokens. Video experiments use Qwen3-VL-2B and Qwen3-VL-8B~\cite{bai2025qwen3} and InternVL3.5-2B~\cite{wang2025internvl3}, with InternVL3.5-8B in \cref{app:scale}, on LongVideoBench with subtitles~\cite{wu2024longvideobench} and the long split of Video-MME~\cite{videomme}, with the pool of GridProbe, 144 uniformly sampled frames on a $12{\times}12$ grid, one frame per block~\cite{gridprobe}. \method stores Qwen3-VL frames at $448{\times}448$ pixels and InternVL3.5 frames at GridProbe's $224{\times}224$ (\cref{app:store}). Image experiments follow AnswerMap~\cite{answermap} with Qwen3-VL-4B on RefCOCOg~\cite{refcocog} and TextVQA~\cite{textvqa}. Prompts follow Periscope for text and GridProbe for video (\cref{app:impl}).

\paragraph{Reads compared.} We compare three reads of the same frozen model, the monolithic read as the reference, grid probing as the emulated read and \method as the native read, with the same prompt and answer read-out and no training. The monolithic read takes the context up to the model's window and truncates the rest, as served models do, and with a prefix cache it encodes each context once and answers every request over that cache. Two training-free methods keep this read and sparsify its prefill, with their published settings. XAttention~\cite{xattention} computes only the attention blocks it scores as important and keeps the prefix cache. Token Sparse Attention~\cite{tokensparse} drops, in each head, the tokens the request attends to least, so its prefill runs for every request. The emulated read is Periscope for text~\cite{periscope}, GridProbe as published for video~\cite{gridprobe}, and the strips of AnswerMap for single images~\cite{answermap}. These three form the grid probing family, and each result names the method that was run.

\paragraph{Metrics.} Accuracy is reported on all questions and on the contexts that fit the window, where the monolithic read sees the whole context. Time to first token (TTFT) is measured for the first request on a context, which includes its encoding (cold), and for later requests on the same context (warm). Time per output token (TPOT) is measured over 100 greedy tokens after the answer, and memory is peak GPU memory. Every timed run uses one A100 80GB, bfloat16, Hugging Face Transformers~\cite{transformers}, and FlashAttention-2~\cite{flashattention2}. \method runs $B{=}8$ sequences per forward and keeps the store on the GPU up to 24GB and in page-locked host memory beyond. Changing only the batch shape of the sequences moves the accuracy of \method by up to three questions in 229 (\cref{app:noise}), so we do not read differences of that size. The probes of \method agree with those of Periscope up to this noise on the local spans (\cref{sec:ablations}), and larger models and a third model family are in \cref{app:scale}.

\paragraph{Reruns on video.} We rerun every read of \cref{tab:video} in one harness, with the released code of GridProbe, the same frames and the same prompt, so the reads differ only in how they read.

\subsection{Noise floor}
\label{app:noise}

The logits leave the model in bfloat16, so near the magnitudes of the answer logits a score moves in steps of 0.125. A batched forward over sequences of different lengths changes the reduction order of the attention kernel and can move a score by a few such steps. \Cref{tab:noise} isolates this effect on Qwen3-4B and En.MC.

\begin{table}[h]
  \centering
  \footnotesize
  \setlength{\tabcolsep}{3pt}
  \begin{tabular}{@{}lccccc@{}}
    \toprule
    \textbf{Comparison} & \textbf{Questions} & \textbf{Median} & \textbf{90th} & \textbf{Max} & \textbf{Changed} \\
    \midrule
    $B{=}8$ against $B{=}1$ & 225 & 0.25 & 0.63 & 1.75 & 6 / 5 \\
    $B{=}1$ against earlier code & 11 & 0 & 0 & 0 & 0 / 0 \\
    $B{=}8$, repeated & 11 & 0 & 0 & 0 & 0 / 0 \\
    \bottomrule
  \end{tabular}
  \caption{\textbf{Effect of the batch shape alone.} Median, 90th percentile, and maximum of the absolute difference of the option scores, and answers that change, select / direct.}
  \label{tab:noise}
\end{table}

Inference itself is deterministic, since running the sequences one at a time reproduces earlier code exactly and a batched run repeats exactly. The only source of variation is the batch shape, which moves a handful of answers out of 225 in both directions, with a net change of at most three questions on either model and either kernel. We therefore treat differences of up to three questions as the noise floor of bfloat16 inference.

\subsection{Confidence intervals}
\label{app:ci}

\begin{table}[h]
  \centering
  \footnotesize
  \setlength{\tabcolsep}{3pt}
  \begin{tabular}{@{}l l c c@{}}
    \toprule
    \textbf{Setting} & \textbf{\method against} & \textbf{Difference} & \textbf{95\% interval} \\
    \midrule
    En.MC, Qwen3-4B & Periscope & $-$1.3 & $-$4.4 to $+$1.7 \\
    229 questions & Monolithic, cached & $+$14.0 & $+$8.3 to $+$19.7 \\
    \midrule
    LVB, InternVL3.5-8B & GridProbe & $+$1.1 & $-$0.3 to $+$2.5 \\
    1{,}202 questions & Monolithic & $+$6.4 & $+$4.1 to $+$8.7 \\
    \bottomrule
  \end{tabular}
  \caption{\textbf{Paired confidence intervals.} Difference in select accuracy (points) between \method and another read on the same questions, with a paired bootstrap interval over questions.}
  \label{tab:ci}
\end{table}

The noise floor bounds the effect of the batch shape, and \cref{tab:ci} adds the sampling error of the questions for one text and one video setting. Against grid probing the interval of the difference contains zero in both settings, so \method matches the accuracy of Periscope and of GridProbe. Against the monolithic read the interval lies well above zero in both settings, so the gain of \method is not an effect of the sample.

\section{Accuracy and cost}
\label{app:accuracy}

This section completes the results of \cref{sec:accuracy} in the setting of \cref{tab:video,tab:text}. It reads the remaining splits of Video-MME, a larger InternVL model and larger text models, a second text benchmark, and token reduction as a further cheaper monolithic read. It then follows the cost of the reads as the input grows.

\subsection{Video-MME, all durations}
\label{app:duration}

\begin{table}[h]
  \centering
  \footnotesize
  \setlength{\tabcolsep}{4pt}
  \begin{tabular}{@{}l cccc@{}}
    \toprule
    & \multicolumn{4}{c}{\textbf{Accuracy} (\%) $\uparrow$} \\
    \cmidrule(l){2-5}
    \textbf{Method} & Short & Medium & Long & Overall \\
    \midrule
    \multicolumn{5}{@{}l}{\emph{InternVL3.5-2B}} \\
    Monolithic read & 61.3 & 49.8 & 44.8 & 52.0 \\
    GridProbe~\cite{gridprobe} & \textbf{65.4} & \textbf{53.9} & \textbf{46.8} & \textbf{55.4} \\
    \oursrow \textbf{\method} & 64.8 & 53.7 & 45.8 & 54.8 \\
    \midrule
    \multicolumn{5}{@{}l}{\emph{InternVL3.5-8B}} \\
    Monolithic read & 72.3 & 60.4 & 52.8 & 61.8 \\
    GridProbe~\cite{gridprobe} & 73.6 & 62.6 & \textbf{54.1} & \textbf{63.4} \\
    \oursrow \textbf{\method} & \textbf{74.6} & \textbf{63.3} & 52.1 & 63.3 \\
    \midrule
    \multicolumn{5}{@{}l}{\emph{Qwen3-VL-2B}} \\
    Monolithic read & \textbf{73.2} & \textbf{58.1} & 47.1 & \textbf{59.5} \\
    GridProbe~\cite{gridprobe} & 70.6 & 54.9 & 46.7 & 57.4 \\
    \oursrow \textbf{\method} & 73.0 & 55.8 & \textbf{47.3} & 58.7 \\
    \midrule
    \multicolumn{5}{@{}l}{\emph{Qwen3-VL-8B}} \\
    Monolithic read & \textbf{80.8} & \textbf{68.3} & \textbf{58.7} & \textbf{69.3} \\
    GridProbe~\cite{gridprobe} & 76.3 & 64.8 & 55.8 & 65.6 \\
    \oursrow \textbf{\method} & 77.7 & 65.7 & 56.2 & 66.5 \\
    \bottomrule
  \end{tabular}
  \caption{\textbf{Video-MME, all durations.} Accuracy (\%) on the 900 questions of each duration split with 144 frames, and over all 2{,}700. The long split of InternVL3.5-2B, Qwen3-VL-2B and Qwen3-VL-8B repeats \cref{tab:video}. Bold marks the best per model.}
  \label{tab:duration}
\end{table}

\Cref{tab:video} uses the long split of Video-MME. \Cref{tab:duration} adds the short and medium splits with the same models, frames and reads, which tests the reading of \cref{sec:accuracy} that the gain of \method follows how much the frames strain the window. InternVL3.5-2B and InternVL3.5-8B write every frame as 256 tokens, so 144 frames nearly fill their window whatever the length of the video. On InternVL3.5-2B \method gains over the monolithic read at every duration, by 3.5, 3.9 and 1.0 points on short, medium and long videos. On InternVL3.5-8B it gains 2.3 and 2.9 points on short and medium videos and stays within 0.7 points on long videos. Over all of Video-MME, \method is 2.8 and 1.5 points above the monolithic read on the two models and within 0.6 points of GridProbe. Qwen3-VL-2B and Qwen3-VL-8B read their 144 frames well inside their window, and there \method stays within 0.8 and 2.8 points of the monolithic read overall and above GridProbe on every split. The gain of \method therefore follows how much the frames strain the window, not the duration of the video. On the short and medium splits \method answers a question in 2.2 to 5.2 seconds, averaged over the three questions on each video, against 3.4 to 11.3 seconds for the monolithic read and 4.2 to 12.4 for GridProbe.

\subsection{Scale and family}
\label{app:scale}
\label{sec:scale}

\begin{table}[h]
  \centering
  \footnotesize
  \setlength{\tabcolsep}{4pt}
  \begin{tabular}{@{}l cc c@{}}
    \toprule
    & \multicolumn{2}{c}{\textbf{Accuracy} (\%) $\uparrow$} & \textbf{Seconds} $\downarrow$ \\
    \cmidrule(lr){2-3}
    \textbf{Method} & LVB & V-MME & LVB \\
    \midrule
    Monolithic read & 56.7 & 52.8 & 6.82 \\
    GridProbe~\cite{gridprobe} & 62.1 & \textbf{54.1} & 14.14 \\
    \oursrow \textbf{\method} & \textbf{63.1} & 52.1 & \textbf{6.44} \\
    \bottomrule
  \end{tabular}
  \caption{\textbf{InternVL3.5-8B on long video.} Accuracy (\%) on LongVideoBench (LVB) and Video-MME long (V-MME) with 144 frames, and seconds per LVB question averaged over the questions on a video. Bold marks the best per column.}
  \label{tab:scale-video}
\end{table}

The gain of \method on a model whose window the frames nearly fill should hold at a larger size. \Cref{tab:video} reads Qwen3-VL at 2B and 8B, and InternVL3.5-8B completes the pair for the InternVL family, whose window 144 frames nearly fill (\cref{tab:scale-video}). On LongVideoBench, \method is 6.4 points above the monolithic read and about a point above GridProbe (\cref{tab:ci}), and it answers a question in 6.4 seconds on average, less than half the time of GridProbe and slightly less than the monolithic read. The gain of \method thus holds at four times the size on LongVideoBench, and \cref{tab:duration} gives the three splits of Video-MME for the same model.

\begin{table}[t]
  \centering
  \small
  \setlength{\tabcolsep}{3pt}
  \begin{tabular}{@{}lcccc@{}}
    \toprule
    & & & \multicolumn{2}{c}{\textbf{\method}} \\
    \cmidrule(l){4-5}
    \textbf{Model} & \textbf{Window} & \textbf{Monolithic} & Select & Direct \\
    \midrule
    \textcolor{gray}{Qwen3-4B} & \textcolor{gray}{262k} & \textcolor{gray}{70.3} & \textcolor{gray}{81.1} & \textcolor{gray}{77.3} \\
    Qwen3-14B & 131k & 72.2 & 83.5 & 79.7 \\
    Qwen3-32B & 131k & 82.3 & 86.1 & 84.8 \\
    \midrule
    \textcolor{gray}{Llama-3.1-8B} & \textcolor{gray}{131k} & \textcolor{gray}{74.7} & \textcolor{gray}{77.2} & \textcolor{gray}{81.0} \\
    Mistral-Nemo-12B & 128k & 49.4 & 67.1 & 55.7 \\
    \bottomrule
  \end{tabular}
  \caption{\textbf{Scale and family on En.MC.} Accuracy (\%) on the contexts inside the window. Gray rows repeat \cref{tab:text}.}
  \label{tab:scale}
\end{table}

On text, \cref{tab:scale} repeats the in-window comparison of \cref{tab:text} on larger models and on a third family. Qwen3-14B and Qwen3-32B are read with the 131k window their model cards recommend, from a native window of 32k, and stair positions exceed 32k only for novels above about 500k tokens. Mistral-Nemo-12B adds a third family, read with its 128k window. The gain of \method over the monolithic read holds at every size and in the third family, and it shrinks as the model reads its own window better. It is largest on Mistral-Nemo-12B, whose monolithic read is weakest, and smallest on Qwen3-32B, the strongest reader of the four. Beyond the window the advantage returns even on Qwen3-32B, where \method keeps 86.2 against 68.0 for the prefix that the monolithic read sees.

\subsection{LongBench v2}
\label{app:longbench}

\begin{table}[t]
  \centering
  \small
  \setlength{\tabcolsep}{3pt}
  \begin{tabular}{@{}lcccc@{}}
    \toprule
    & \multicolumn{2}{c}{\textbf{Qwen3-4B}} & \multicolumn{2}{c}{\textbf{Llama-3.1-8B}} \\
    \cmidrule(lr){2-3}\cmidrule(l){4-5}
    \textbf{Method} & Accuracy & Tokens & Accuracy & Tokens \\
    \midrule
    Monolithic read & 36.2 & 86k & 27.6 & 73k \\
    Periscope~\cite{periscope} & 37.0 / 34.7 & 186k & 30.6 / 28.6 & 175k \\
    \oursrow \textbf{\method} & 35.9 / 34.9 & 83k & 29.6 / 28.6 & 78k \\
    \bottomrule
  \end{tabular}
  \caption{\textbf{LongBench v2, short and medium splits.} Accuracy (\%) select / direct on 395 questions, and context tokens processed per question.}
  \label{tab:longbench}
\end{table}

A second text benchmark checks the result of \cref{tab:text}. LongBench v2 gives almost every question its own context, so the reuse of the encoding plays no role and the benchmark tests accuracy and work alone (\cref{tab:longbench}). \method stays within about a point of Periscope on both read-outs while processing less than half of its tokens, so even a single question on a context takes \method less time than Periscope. Against the monolithic read, \method is ahead with Llama-3.1-8B and within about a point with Qwen3-4B, whose window holds almost every context of the benchmark, which matches the main result that probing helps where the context strains the model.

\subsection{Other ways to make the monolithic read cheaper}
\label{app:tokenred}

\begin{table}[h]
  \centering
  \footnotesize
  \setlength{\tabcolsep}{3pt}
  \begin{tabular}{@{}l c cc cc@{}}
    \toprule
    & & \multicolumn{2}{c}{\textbf{Accuracy} (\%) $\uparrow$} & \multicolumn{2}{c}{\textbf{TTFT} (s) $\downarrow$} \\
    \cmidrule(lr){3-4}\cmidrule(l){5-6}
    \textbf{Method} & \textbf{Tokens} & V-MME & LVB & First & Later \\
    \midrule
    Monolithic read & 100\% & 47.1 & 56.2 & 6.40 & 6.40 \\
    VidCom2~\cite{vidcom2} & 25\% & 46.3 & \textbf{56.4} & 4.62 & 4.62 \\
    FastVID~\cite{fastvid} & 25\% & 45.7 & 55.6 & \textbf{4.20} & 4.20 \\
    VisionZip~\cite{visionzip} & 25\% & 43.9 & 56.3 & 4.42 & 4.42 \\
    HoliTom~\cite{holitom} & 25\% & 44.6 & \textbf{56.4} & 4.82 & 4.82 \\
    GridProbe~\cite{gridprobe} & -- & 46.7 & 55.1 & 4.53 & 4.53 \\
    \oursrow \textbf{\method} & -- & \textbf{47.3} & 56.3 & 5.05 & \textbf{2.60} \\
    \bottomrule
  \end{tabular}
  \caption{\textbf{Token reduction next to the sparse prefill, Qwen3-VL-2B.} Accuracy (\%) on Video-MME long (V-MME) and LongVideoBench (LVB) with 144 frames. Tokens is the share of the video tokens that the language model reads, and TTFT is the time to first token for the first and a later question on a video, as in \cref{tab:video}. Bold marks the best per column.}
  \label{tab:tokenred}
\end{table}

\Cref{tab:video} includes XAttention, a training-free sparse prefill of the monolithic read, and \cref{sec:accuracy} finds that XAttention saves little time on video, because the vision encoder dominates the cost of a video read. Token reduction is the other common training-free way to make the monolithic read cheaper. It drops video tokens between the vision encoder and the language model. \Cref{tab:tokenred} runs four such methods in the setting of \cref{tab:video} on Qwen3-VL-2B, VidCom2~\cite{vidcom2}, FastVID~\cite{fastvid}, VisionZip~\cite{visionzip} and HoliTom~\cite{holitom}, through the implementations for Qwen3-VL released with VidCom2, with 144 frames and 25\% of the video tokens, the default of that code. On LongVideoBench all four stay within 0.6 points of the full read, while on Video-MME long they lose 0.8 to 3.2 points. \method is more accurate than all four on Video-MME long and within 0.1 points of the best of them on LongVideoBench. Like XAttention, these methods still run the vision encoder over every frame for every question. With a quarter of the video tokens they take 4.2 to 4.8 seconds on every question, against 6.4 for the full read. \method takes 5.05 seconds on a first question about a video and 2.60 seconds on a later one, since it reuses its encoding.

\subsection{Cost against frames}
\label{app:scaling}

\begin{figure}[h]
  \centering
  \includegraphics[width=\linewidth]{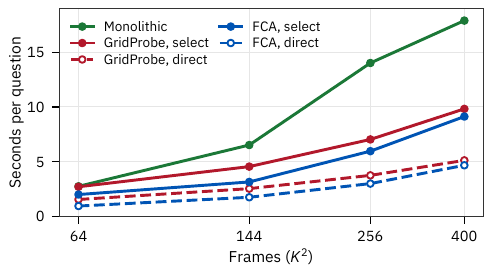}
  \caption{\textbf{Cost against frames.} Seconds per question for Qwen3-VL-2B on 20 LongVideoBench questions as the pool grows from $8{\times}8$ to $20{\times}20$ frames, averaged over the questions on a video, with the store of \method at 448 pixels. \Cref{app:scaling} gives the TFLOPs.}
  \label{fig:scaling}
\end{figure}

\begin{figure}[h]
  \centering
  \includegraphics[width=0.7\linewidth]{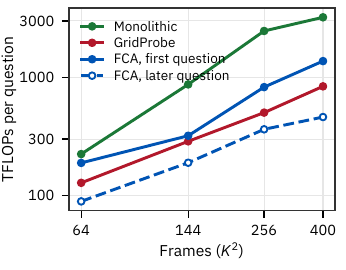}
  \caption{\textbf{TFLOPs against frames.} Qwen3-VL-2B on the 20 LongVideoBench questions of \cref{fig:scaling}, TFLOPs of the language model per question.}
  \label{fig:scaling-tflops}
\end{figure}

The times of \cref{tab:video} are measured at 144 frames, and \cref{fig:scaling} follows them as the number of frames grows. The cost of \method grows more slowly with the input than the cost of the other reads. From 64 to 400 frames \method stays the fastest read, while the monolithic read becomes the slowest from 144 frames on, and the same holds for text (\cref{app:costlength}). \Cref{fig:scaling-tflops} gives the TFLOPs of the reads of \cref{fig:scaling}, by the count of GridProbe for the language model. The monolithic read grows fastest, since it reads every frame at near-native resolution. \method costs about as much as GridProbe at every pool size while storing frames at twice the resolution, and a later question on a video costs about half of a first one.

\subsection{Store resolution}
\label{app:store}

The cost of a video read also depends on the resolution at which the frames are stored. Because \method encodes every frame once, a larger store costs one encoding pass rather than $2K$ probe passes. \Cref{tab:store} compares the store of Qwen3-VL-2B at the probe resolution of GridProbe and at twice that side. The larger store raises both read-outs on both benchmarks, most on Video-MME, for about 15\% more compute. The extra cost falls on the first question about a video, while a later question takes almost the same time, so the higher resolution is nearly free over a conversation.

\begin{table}[h]
  \centering
  \footnotesize
  \begin{subtable}[b]{0.32\linewidth}
    \centering
    \setlength{\tabcolsep}{2.5pt}
    \begin{tabular}{@{}lcc@{}}
      \toprule
      \textbf{Store} & \textbf{LVB} & \textbf{V-MME} \\
      \midrule
      224 & 55.4 & 44.3 \\
      \oursrow 448 & \textbf{56.3} & \textbf{47.3} \\
      \bottomrule
    \end{tabular}
    \caption{Select acc.}
    \label{tab:store-acc}
  \end{subtable}\hfill
  \begin{subtable}[b]{0.32\linewidth}
    \centering
    \setlength{\tabcolsep}{2.5pt}
    \begin{tabular}{@{}lcc@{}}
      \toprule
      \textbf{Store} & \textbf{LVB} & \textbf{V-MME} \\
      \midrule
      224 & \textbf{276} & \textbf{194} \\
      \oursrow 448 & 321 & 220 \\
      \bottomrule
    \end{tabular}
    \caption{TFLOPs per question.}
    \label{tab:store-flops}
  \end{subtable}\hfill
  \begin{subtable}[b]{0.32\linewidth}
    \centering
    \setlength{\tabcolsep}{2.5pt}
    \begin{tabular}{@{}lcc@{}}
      \toprule
      \textbf{Store} & \textbf{Cold} & \textbf{Warm} \\
      \midrule
      224 & \textbf{3.41} & \textbf{2.43} \\
      \oursrow 448 & 5.05 & 2.60 \\
      \bottomrule
    \end{tabular}
    \caption{TTFT on LVB (s).}
    \label{tab:store-time}
  \end{subtable}
  \caption{\textbf{Resolution of the video store.} Qwen3-VL-2B, 144 frames, $K{=}12$, store side in pixels. GridProbe probes at 224.}
  \label{tab:store}
\end{table}

\section{Reach beyond the window}
\label{app:reach}

This section completes the needle tests of \cref{sec:reach}. It gives the details of the video and text needles, and it adds Periscope to the two reads of \cref{tab:needle}.

\paragraph{Video needles.} The needle test checks the reach on video first. An image from TextVQA~\cite{textvqa} replaces one frame of a Video-MME long video at one of 11 depths, and the model answers a question about the image with four options, with about 33 needles per length (\cref{tab:vneedle}). Accuracy counts only the needles that each model answers when shown the image alone, and the select read of both probing methods reads at most 144 frames. The trend follows how each model fits a long video into its window. InternVL3.5-2B writes every frame as 256 tokens at any resolution, so its window of 40{,}960 tokens holds at most 158 frames. Its monolithic read therefore samples 144 frames uniformly, the standard practice for long video, which keeps it fast at every length but leaves most of a long video unread, and the sampled frames contain the needle less and less often. Its accuracy falls by about 17 points from 144 to 2{,}025 frames, while \method stays within one or two needles of GridProbe at every length and answers in about 60\% of the time of GridProbe at the two longest lengths. Qwen3-VL-2B reads every frame by shrinking each one, up to 1{,}024 frames, the most that fits in memory, so its monolithic read degrades more gently, and \method stays above it at every length. One needle moves the accuracy by about 3 points, so the small gaps between \method and GridProbe are noise, while the gap to the monolithic read of InternVL3.5-2B is not.

\paragraph{Text needles.} Text gives the same picture at a larger scale. Following RULER~\cite{ruler}, a sentence that pairs a word with a seven-digit number is inserted at 11 depths into the En.MC novels, cut to 32k to 512k tokens, with three needles per length and depth, and the generated answer is scored by string match (\cref{tab:needle,fig:needle}). \Cref{fig:needle} gives the accuracy by length and depth. Each needle is the first request on its context, so the times of \cref{tab:needle} include the encoding.

\subsection{Text needles, all three reads}
\label{app:niah-periscope}

\begin{table}[h]
  \centering
  \footnotesize
  \setlength{\tabcolsep}{3.5pt}
  \begin{tabular}{@{}l cc cc c@{}}
    \toprule
    & \multicolumn{2}{c}{\textbf{Match} (\%) $\uparrow$} & \multicolumn{2}{c}{\textbf{Seconds} $\downarrow$} & \textbf{Mem.} \\
    \cmidrule(lr){2-3}\cmidrule(lr){4-5}
    \textbf{Method} & In-window & Beyond & In-window & All & (GB) \\
    \midrule
    \multicolumn{6}{@{}l}{\emph{Qwen3-4B, window 262k}} \\
    Monolithic read & \textbf{97.7} & 45.5 & 39.3 & 54.2$^{*}$ & 49.3 \\
    Periscope~\cite{periscope} & 95.5 & \textbf{100.0} & 16.5 & 28.9 & 13.3 \\
    \oursrow \textbf{\method} & 95.5 & \textbf{100.0} & 12.6 & 23.4 & 37.9 \\
    \midrule
    \multicolumn{6}{@{}l}{\emph{Llama-3.1-8B, window 131k}} \\
    Monolithic read & \textbf{97.0} & 36.4 & 15.3 & 22.2$^{*}$ & 37.2 \\
    Periscope~\cite{periscope} & 93.9 & \textbf{98.5} & 13.5 & 37.6 & 24.0 \\
    \oursrow \textbf{\method} & 93.9 & \textbf{98.5} & 8.0 & 26.5 & 42.7 \\
    \bottomrule
  \end{tabular}
  \caption{\textbf{Text needles, all three reads.} String match (\%) of the 165 needles of \cref{tab:needle} inside and beyond the window, mean seconds per needle, each the first request on its context, and peak GPU memory. Bold marks the higher accuracy per model. $^{*}$Reads only the prefix that fits the window.}
  \label{tab:niah-periscope}
\end{table}

\Cref{tab:needle} compares \method with the monolithic read on the text needles, and \cref{tab:niah-periscope} adds Periscope, so the needle test covers all three reads of \cref{sec:experiments}. On both models Periscope and \method find the same share of needles inside the window and beyond it, so native attention keeps the full reach of grid probing. Every needle is the first request on its own context, so neither read reuses an encoding, and \method still answers 1.2 to 1.7 times faster than Periscope. The memory follows the trade-off of \cref{sec:exec}. Periscope holds one span at a time, 13 to 24 GB, and \method keeps the encoding of the whole context for later questions, 38 to 43 GB.

\section{Many questions on one input}
\label{app:many}

This section completes \cref{sec:openended}. It gives the details of the conversations of \cref{tab:multiturn}, the read for open-ended questions, and the conversations with long turns of \cref{tab:append}.

\paragraph{Conversations.} On video, a later question reads the stored frames instead of encoding them again (\cref{tab:video}), so it takes \method a quarter to two thirds of the time of GridProbe, which encodes its probes again for every question. The direct read-out of \method skips the focused pass and answers a later question in under two seconds, while staying above the direct read-out of GridProbe. A user who reads a document with a model asks a sequence of questions about it, and every later question is a request over the same context. SCBench~\cite{scbench} tests this setting with sessions of up to five questions over one InfiniteBench novel, open-ended (En.QA) or with four options (En.MC). Each turn carries the earlier questions and their reference answers in its request, answers are generated greedily up to 40 tokens, and the monolithic read is cut in the middle to fit its window, as in SCBench. Open-ended questions are read as \cref{app:openended} describes. Over a conversation (\cref{tab:multiturn}), \method keeps the accuracy of Periscope on open questions and stays within about three points of Periscope on multiple choice, well above the monolithic read on both, and it answers each later turn five to seven times faster than Periscope. A cached monolithic read remains the fastest per turn, and \method brings the accuracy of probing to a later turn in about five seconds.

\subsection{Open-ended questions and conversations}
\label{app:openended}

An open-ended question has no answer options to score, so each probe scores relevance. After the question, every span sequence asks whether the text above contains the information needed to answer it, and the probe score is the log-odds of Yes against No~\cite{periscope}, with Yes in the role of an option and No in the role of the abstain option of \cref{eq:probe}. The scores are pooled into the evidence map as in \cref{eq:pool}, and the select read-out reads the $K$ blocks with the highest cells again, in their original order, with the question, and generates the answer. The direct read-out does not apply, since there are no options to choose among.

In a conversation, every turn is a new request over the same store. The request of a turn carries the previous questions and answers followed by the new question, so it grows with every turn, while the store never changes and the context is never encoded again. Each turn runs its full request in all $2K$ sequences and then discards it. This suits conversations with short turns, such as those of SCBench (\cref{sec:openended}), where the history stays at a few hundred tokens. With long turns, the request itself would grow beyond the range that the model reads well and would be processed $2K$ times per turn. Long turns can instead enter the store as new local spans, encoded once each, which \cref{app:append} tests.

\subsection{Conversations with long turns}
\label{app:append}

This section measures the encoding of new input (\cref{sec:exec}). Each of 8 conversations has 16 turns. Every turn adds 32k tokens of the En.MC novels with one needle sentence at a random place, as in the needle test, and asks for the number of a needle from a turn chosen uniformly among the turns so far. The answer is generated and scored by string match. \method keeps a grid with a fixed number of columns, $K{=}33$, and every turn appends its blocks as new rows. Only the rows that receive new blocks are encoded, and a row that was partly filled is encoded again as a whole, so every row is always one causal pass. The turn's question is then answered as any request, by one sequence per row and per column and the select read-out, and the request holds only the question. The monolithic read keeps the most recent turns that fit its window, and Periscope probes the whole conversation again at every turn.

The conversation grows to 524k tokens, four times the window of Llama-3.1-8B and twice that of Qwen3-4B, and the accuracy of \method does not fall with the turn (\cref{tab:append}). \method finds a needle from ten or more turns ago as often as a needle of the current turn, and its evidence map ranks the block of the needle first in the median case. The monolithic read loses every turn that no longer fits its window, so its accuracy falls with the turn, and it finds no needle from ten or more turns ago. Once the conversation exceeds the window, the turns that the monolithic read keeps change at every turn, so it also encodes them again at every turn. With Qwen3-4B, Periscope finds every needle and \method 95.3\% of them, while the time per turn of Periscope grows with the length of the conversation, to about five times the time of \method in the last turns. With Llama-3.1-8B, Periscope finds 96.9\% of the needles, against 98.4\% for \method, and takes 51.8 seconds per turn on average, against 10.7. \method spends about 3 seconds on encoding a new turn of 32k tokens from the fifth turn to the last, so the cost of new input does not depend on what is already stored. The time of a whole turn of \method grows from about 9 to about 14 seconds over the conversation, since every new row adds one sequence to the request.

\section{Evidence localization}
\label{app:local}

This section completes \cref{sec:images}. The video needles place the evidence in one known frame, so the map of \method can be scored against that frame, and three examples then show the maps themselves.

\subsection{Localization on the video needles}
\label{app:vloc}

\begin{table*}[h]
  \centering
  \footnotesize
  \begin{subtable}[b]{0.49\linewidth}
    \centering
    \setlength{\tabcolsep}{4pt}
    \begin{tabular}{@{}l ccccc@{}}
      \toprule
      \textbf{Frames} & 144 & 256 & 576 & 1{,}024 & 2{,}025 \\
      \midrule
      \multicolumn{6}{@{}l}{\emph{InternVL3.5-2B, 29 needles}} \\
      GridProbe~\cite{gridprobe} & \textbf{69.0} & 55.2 & \textbf{65.5} & 51.7 & \textbf{51.7} \\
      \oursrow \textbf{\method} & 62.1 & \textbf{62.1} & 62.1 & \textbf{55.2} & \textbf{51.7} \\
      \midrule
      \multicolumn{6}{@{}l}{\emph{Qwen3-VL-2B, 33 needles}} \\
      GridProbe~\cite{gridprobe} & \textbf{75.8} & \textbf{75.8} & 69.7 & 69.7 & 63.6 \\
      \oursrow \textbf{\method} & \textbf{75.8} & \textbf{75.8} & \textbf{75.8} & \textbf{72.7} & \textbf{69.7} \\
      \bottomrule
    \end{tabular}
    \caption{Peak of the map on the needle frame (\%)}
    \label{tab:vloc-peak}
  \end{subtable}\hfill
  \begin{subtable}[b]{0.49\linewidth}
    \centering
    \setlength{\tabcolsep}{4pt}
    \begin{tabular}{@{}l ccccc@{}}
      \toprule
      \textbf{Frames} & 144 & 256 & 576 & 1{,}024 & 2{,}025 \\
      \midrule
      \multicolumn{6}{@{}l}{\emph{InternVL3.5-2B, 29 needles}} \\
      GridProbe~\cite{gridprobe} & \textbf{89.7} & \textbf{86.2} & 82.8 & \textbf{79.3} & \textbf{75.9} \\
      \oursrow \textbf{\method} & \textbf{89.7} & \textbf{86.2} & \textbf{86.2} & 75.9 & 69.0 \\
      \midrule
      \multicolumn{6}{@{}l}{\emph{Qwen3-VL-2B, 33 needles}} \\
      GridProbe~\cite{gridprobe} & \textbf{93.9} & \textbf{87.9} & \textbf{87.9} & \textbf{84.8} & 81.8 \\
      \oursrow \textbf{\method} & 87.9 & \textbf{87.9} & 81.8 & \textbf{84.8} & \textbf{84.8} \\
      \bottomrule
    \end{tabular}
    \caption{Needle frame kept by the select read (\%)}
    \label{tab:vloc-kept}
  \end{subtable}
  \caption{\textbf{Localization on the video needles.} Share of the needles each model answers from the image alone, 29 of 33 for InternVL3.5-2B and all 33 for Qwen3-VL-2B. The median rank of the needle frame is 1 at every length for both methods and both models. Bold marks the higher value per model.}
  \label{tab:vloc}
\end{table*}

The video needle test of \cref{sec:reach} places the evidence in one known frame, so the map can be scored against that frame (\cref{tab:vloc}). On both models and at every length from 144 to 2{,}025 frames, the map of \method ranks the needle frame first in the median case, as the map of GridProbe does. With InternVL3.5-2B, the peak of the map of \method is the needle frame for 52 to 62\% of the needles, against 52 to 69\% for GridProbe, and its select read keeps the needle frame for 69 to 90\%, against 76 to 90\%. With Qwen3-VL-2B, the peak of the map of \method is the needle frame for 70 to 76\% of the needles, against 64 to 76\% for GridProbe, and its select read keeps the needle frame for 82 to 88\%, against 82 to 94\%. At every length and on both models, the two maps differ by at most two needles on either measure, so the native read localizes the evidence as well as the emulated read.

\subsection{Evidence maps}
\label{app:maps}

\begin{figure}[h]
  \centering
  \includegraphics[width=\linewidth]{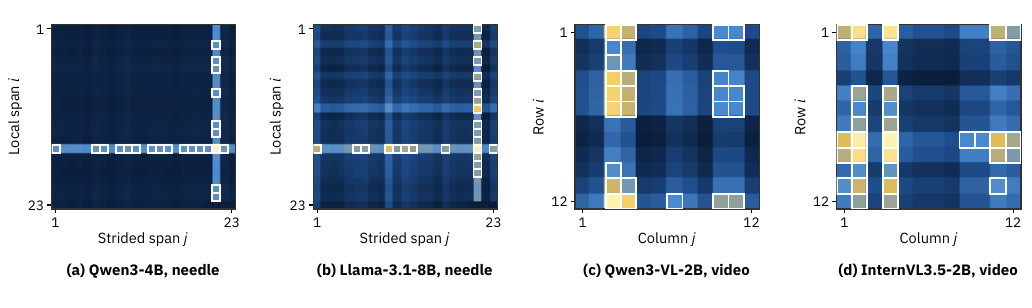}
  \caption{\textbf{Evidence maps of \method.} (a, b) Map of a needle question, $M[i,j]$ the sum of the Yes log-odds of local span $\Loc_i$ and strided span $\Str_j$, cell $(i, j)$ being block $23(i{-}1){+}j$. (c, d) Map of a LongVideoBench question, the product of the row and column confidences, cell $(i, j)$ being frame $12(i{-}1){+}j$. Outlined cells are read again by the select read-out.}
  \label{fig:maps}
\end{figure}

\Cref{fig:maps} shows the evidence map of \method on a needle question and on two video questions. In the needle example, a sentence with a seven-digit number sits at 70\% of a 262k-token haystack, in local span $\Loc_{16}$ of a $23{\times}23$ grid. The maps of both text models peak at the cell of that block, because only the probes of $\Loc_{16}$ and of the strided span $\Str_{21}$ answer Yes, and the select read-out reads the outlined cells and returns the number. In the two LongVideoBench examples, the outlined frames are the ones the map selects for the focused pass, about a fifth of the 144 frames, and the focused pass answers both questions correctly.

\section{Ablations}
\label{app:abl}

This section completes \cref{sec:ablations}. It compares the probes of \method with those of Periscope one by one, tests the length of the select read, and ablates the remaining parts of \method.

\subsection{Agreement with Periscope}
\label{app:agreement}

\Cref{tab:agreement} compares every probe score of \method on En.MC with the score that Periscope~\cite{periscope} computes for the same span and option, for the two schedules that encode local spans. The last row is the difference produced by changing only the batch shape (\cref{app:noise}). The two reads also choose much of the same evidence. On En.MC with Qwen3-4B, 64\% of the blocks that the select read of \method reads are also read by the select read of Periscope, and the two maps peak at the same cell for 43\% of the questions.

\begin{figure}[h]
  \centering
  \includegraphics[width=\linewidth]{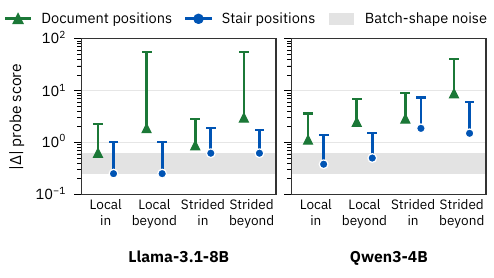}
  \caption{\textbf{Agreement with Periscope.} Absolute difference between the probe scores of \method and of Periscope on En.MC, median (marker) and 90th percentile (bar end). Gray: changing only the batch shape.}
  \label{fig:schedule}
\end{figure}

\begin{table}[h]
  \centering
  \footnotesize
  \setlength{\tabcolsep}{2.2pt}
  \begin{tabular}{@{}llcccc@{}}
    \toprule
    & & \multicolumn{2}{c}{\textbf{Local spans}} & \multicolumn{2}{c}{\textbf{Strided spans}} \\
    \cmidrule(lr){3-4}\cmidrule(l){5-6}
    \textbf{Model} & \textbf{Positions} & In & Beyond & In & Beyond \\
    \midrule
    Llama-3.1-8B & Document & 0.63 (2.3) & 1.88 (55.7) & 0.88 (2.8) & 3.00 (54.8) \\
    & Stair & 0.25 (1.0) & 0.25 (1.0) & 0.62 (1.9) & 0.62 (1.8) \\
    \midrule
    Qwen3-4B & Document & 1.13 (3.6) & 2.50 (7.0) & 2.88 (8.9) & 8.88 (40.3) \\
    & Stair & 0.38 (1.4) & 0.50 (1.5) & 1.87 (7.4) & 1.50 (6.1) \\
    \midrule
    \multicolumn{2}{@{}l}{Batch shape only} & \multicolumn{4}{c}{0.25 (0.6)} \\
    \bottomrule
  \end{tabular}
  \caption{\textbf{Agreement with Periscope.} Absolute difference between the probe scores of \method and of Periscope, median (90th percentile), inside and beyond the window.}
  \label{tab:agreement}
\end{table}

\subsection{Length of the select read}
\label{app:budget}

The select read-out of \method reads the top $K$ blocks of its map again. GridProbe~\cite{gridprobe} instead sets the length of its focused pass from the shape of its map, $\lceil n/(1{+}3(|\gamma|{+}0.5\max(0,\kappa)))\rceil$ of $n$ cells with skewness $\gamma$ and excess kurtosis $\kappa$ of the cell confidences, and at least two. \Cref{tab:budget} applies this rule to the blocks of \method on En.MC, with the confidence of a cell the product of the top answer probabilities of its row and column probes, and at most as many blocks as the window holds. On video the rule keeps about a third of the frames (\cref{sec:video}), but on text maps are flatter and the rule reads several times more blocks than $K$. The longer read does not help. It lowers the accuracy of Qwen3-4B and leaves Llama-3.1-8B unchanged, so \method keeps $K$ blocks and a short select read.

\begin{table}[h]
  \centering
  \footnotesize
  \begin{subtable}[b]{0.48\linewidth}
    \centering
    \setlength{\tabcolsep}{3pt}
    \begin{tabular}{@{}lcc@{}}
      \toprule
      \textbf{Budget} & \textbf{Qwen} & \textbf{Llama} \\
      \midrule
      \oursrow $K$ blocks (default) & \textbf{79.0} & \textbf{79.0} \\
      Skew rule~\cite{gridprobe} & 74.2 & \textbf{79.0} \\
      \bottomrule
    \end{tabular}
    \caption{Select acc.}
    \label{tab:budget-acc}
  \end{subtable}\hfill
  \begin{subtable}[b]{0.48\linewidth}
    \centering
    \setlength{\tabcolsep}{3pt}
    \begin{tabular}{@{}lcc@{}}
      \toprule
      \textbf{Budget} & \textbf{Qwen} & \textbf{Llama} \\
      \midrule
      \oursrow $K$ blocks (default) & \textbf{19.2} & \textbf{19.2} \\
      Skew rule~\cite{gridprobe} & 136.7 & 87.4 \\
      \bottomrule
    \end{tabular}
    \caption{Blocks read.}
    \label{tab:budget-blocks}
  \end{subtable}
  \caption{\textbf{Number of blocks of the select read.} All 229 En.MC questions. (a) Select accuracy (\%). (b) Mean number of blocks of 500 tokens read again. Qwen is Qwen3-4B and Llama is Llama-3.1-8B.}
  \label{tab:budget}
\end{table}

\subsection{Further ablations}
\label{app:ablations}

\begin{table*}[t]
  \centering
  \footnotesize
  \begin{subtable}[b]{0.3\linewidth}
    \centering
    \setlength{\tabcolsep}{3pt}
    \begin{tabular}{@{}lcc@{}}
      \toprule
      \textbf{Read} & \textbf{Qwen} & \textbf{Llama} \\
      \midrule
      One query, as PCW~\cite{pcw} & 24.9 & 24.9 \\
      \oursrow Span queries & \textbf{52.8} & \textbf{36.7} \\
      \bottomrule
    \end{tabular}
    \caption{Read, direct acc.}
    \label{tab:abl-read}
  \end{subtable}\hfill
  \begin{subtable}[b]{0.2\linewidth}
    \centering
    \setlength{\tabcolsep}{3pt}
    \begin{tabular}{@{}lcc@{}}
      \toprule
      \textbf{Tokens} & \textbf{Qwen} & \textbf{Llama} \\
      \midrule
      250 & 77.3 & \textbf{81.2} \\
      \oursrow 500 (default) & \textbf{79.0} & 79.0 \\
      1000 & \underline{78.2} & \underline{80.8} \\
      \bottomrule
    \end{tabular}
    \caption{Block size, select acc.}
    \label{tab:abl-block}
  \end{subtable}\hfill
  \begin{subtable}[b]{0.28\linewidth}
    \centering
    \setlength{\tabcolsep}{3pt}
    \begin{tabular}{@{}lcc@{}}
      \toprule
      \textbf{Sequences} & \textbf{Qwen} & \textbf{Llama} \\
      \midrule
      \oursrow $B{=}8$ (default) & 14.8 / 2.9 & 10.4 / 1.7 \\
      $B{=}1$ & 15.7 / 3.6 & 10.9 / 2.1 \\
      Pageable store & 15.5 / 3.3 & 10.4 / 1.6 \\
      \bottomrule
    \end{tabular}
    \caption{Execution, TTFT cold / warm (s).}
    \label{tab:abl-exec}
  \end{subtable}\hfill
  \begin{subtable}[b]{0.2\linewidth}
    \centering
    \setlength{\tabcolsep}{3pt}
    \begin{tabular}{@{}lcc@{}}
      \toprule
      \textbf{Spans} & \textbf{Qwen} & \textbf{Llama} \\
      \midrule
      Local only & \textbf{79.0} & \underline{78.6} \\
      Strided only & 69.9 & 75.1 \\
      \oursrow Both (default) & \underline{75.1} & \textbf{80.3} \\
      \bottomrule
    \end{tabular}
    \caption{Span families, direct acc.}
    \label{tab:abl-spans}
  \end{subtable}
  \caption{\textbf{Further ablations on En.MC.} All 229 questions, with accuracy in \% and times inside the window. Panel (a) reads one store of blocks encoded alone at the block start, and panel (d) reads one span family alone, in direct accuracy. Qwen is Qwen3-4B, Llama is Llama-3.1-8B, and the default store is page-locked.}
  \label{tab:ablation-more}
\end{table*}

The two span families play different roles. On Llama-3.1-8B neither family alone matches the pooled read, so both are needed (\cref{tab:abl-spans}). On Qwen3-4B the local spans alone already beat the pooled read, while Periscope gains from pooling the same two families, which points to the strided spans of \method as the source of the loss. Counting answers that differ from Periscope confirms this (\cref{tab:abl-approx}). The local spans of \method agree with the local spans of Periscope about as closely as a change of batch shape alone allows (\cref{app:noise}), while the strided spans, whose blocks were encoded within their local spans, change several times more answers. The strided spans are therefore the one approximation of \method, as \cref{sec:pattern} predicts. The select read-out reads the chosen blocks again and recovers most of this gap. Encoding every strided span in its own pass as well removes the approximation at twice the encoding cost and closes most of the remaining gap of the direct read-out on Qwen3-4B, but moves the select read-out by at most about a point on either model (\cref{tab:abl-enc}), so the default keeps one encoding pass per block. Probe by probe (\cref{fig:schedule}), the local probe scores of Llama-3.1-8B under stair positions differ from those of Periscope by no more than a change of batch shape, inside and beyond the window alike. Under document positions the same differences grow by more than ten times beyond the window, and the strided probes differ more, most on Qwen3-4B. \Cref{app:agreement} lists the values.

How a block is encoded and where it is placed matter more than any other choice. A block encoded alone starts without the text before it, and encoding each block within its local span instead adds several points of direct accuracy on Qwen3-4B and far more on Llama-3.1-8B (\cref{tab:abl-enc}). Positions decide whether the store can be read past the window. Document positions keep the order of the blocks but place the late spans of long novels beyond the window, which costs Llama-3.1-8B about a third of its direct accuracy, and costs Qwen3-4B, whose 262k window holds most novels, only a few points (\cref{tab:abl-pos}). Placing every block at the start instead keeps every position small but loses the order of the blocks within a span, and loses far more accuracy (\cref{tab:abl-read,tab:abl-enc}). Stair positions are the only assignment that keeps the order of every span and stays below $2\sqrt{sc}$.

\Cref{tab:ablation-more} ablates three further parts of \method on En.MC, how a request reads the store, the block size and the execution.

The factorized read is what lets a frozen model use a store of separately encoded blocks. On a store of blocks encoded alone at the block start, one query over every block, as parallel encoding~\cite{pcw} reads it, collapses to chance and answers the same letter for almost every question, while the $2K$ span sequences of \method over the same store score well above chance, most on Qwen3-4B (\cref{tab:abl-read}). Methods that let one query read many separately encoded blocks fine-tune the model for it~\cite{blockattention} or recalibrate its attention~\cite{ape}. Hundreds of blocks at the same positions give one query an attention pattern that the frozen model never saw in training, while each sequence of \method reads one span as contiguous text at trained positions.

The block size is not a sensitive choice. Periscope loses several points on LongBench v2 when its blocks double from 500 to 1{,}000 tokens~\cite{periscope}, while the select accuracy of \method stays within about two points between blocks of 250 and 1{,}000 tokens on both models (\cref{tab:abl-block}). The execution choices change only time. Running the sequences one at a time changes about as many answers as the batch shape alone (\cref{app:noise}) and makes a later request slower, and a pageable host store changes no answer and adds a fraction of a second where the store leaves the GPU.

\section{Scope and choice of read}
\label{app:limits}

\Cref{sec:exec} compares emulated and native attention, and three points complete that comparison. First, \method applies to every model whose layers keep a per-token key-value cache, which covers all the language and vision-language models of \cref{sec:experiments}. Second, the three reads serve different inputs. The monolithic read with a prefix cache suits an input that fits the window with room to spare, grid probing suits closed models and small memory, and \method suits inputs near or beyond the window and several questions about one input. Third, each probe reads one span, so evidence that is split across two blocks with no span in common is brought together by the select read.

\end{document}